\documentclass{article}

\PassOptionsToPackage{numbers, compress}{natbib}
\usepackage[final]{neurips_2021}
\usepackage{xr-hyper}       
\usepackage[utf8]{inputenc} 
\usepackage[T1]{fontenc}    
\usepackage{hyperref}       
\usepackage{url}            
\usepackage{booktabs}       
\usepackage{amsfonts}       
\usepackage{nicefrac}       
\usepackage{microtype}      
\usepackage{xcolor}         

\title{On Data-centric Myths}

\author{%
  Antonia Marcu \\
  Vision, Learning and Control\\
  University of Southampton\\
  \texttt{am1g15@soton.ac.uk} \\
   \And
    Adam Pr\"ugel-Bennett\\
    Vision, Learning and Control\\
    University of Southampton\\
    \texttt{apb@ecs.soton.ac.uk} \\
}

\usepackage{todonotes}

\usepackage{algorithm}
\usepackage{algorithmic}

\usepackage{tabularx}
\usepackage{tabulary}
\usepackage{multirow}
\newcolumntype{Y}{>{\centering\arraybackslash}X}

\usepackage{subcaption}

\usepackage{amssymb}

\usepackage{bm}
\usepackage{chngcntr}
\usepackage{float}
\usepackage{stfloats} 

\usepackage{xspace}
\newcommand{\fmix}{FMix\xspace}

\newcommand{\iocc}{iOcclusion\xspace}
\newcommand{\cutocc}{CutOcclusion\xspace}
\newcommand{\mixup}{MixUp\xspace}
\newcommand{\cutmix}{CutMix\xspace}
\newcommand{\cutout}{CutOut\xspace}

\newcommand{\cifar}[1]{CIFAR-{#1}\xspace}

\newcommand{\imagenet}{ImageNet\xspace}

\makeatletter
\newcommand*{\addFileDependency}[1]{
  \typeout{(#1)}
  \@addtofilelist{#1}
  \IfFileExists{#1}{}{\typeout{No file #1.}}
}
\makeatother

\newcommand*{\myexternaldocument}[1]{
    \externaldocument{#1}
    \addFileDependency{#1.tex}
    \addFileDependency{#1.aux}
}
\myexternaldocument{Supplementary}
\usepackage{cleveref}

\usepackage{ifthen}
\newcommand{\linerange}[2]{%
\ifthenelse{\equal{\getrefnumber{#1}}{\getrefnumber{#2}}}{%
\ref{#1}%
}{%
\ref{#1}--\ref{#2}%
}%
}

\begin{document}

\maketitle

\begin{abstract}

    The community lacks theory-informed guidelines for building good data sets. We analyse theoretical directions relating to what aspects of the data matter and conclude that the intuitions derived from the existing literature are incorrect and misleading. 
    Using empirical counter-examples, we show that 1) data dimension should not necessarily be minimised and 2) when manipulating data, preserving the distribution is inessential.
    This calls for a more data-aware theoretical understanding. Although not explored in this work, we propose the study of the impact of data modification on learned representations as a promising research direction. 
    
\end{abstract}

\section{Motivation}
\label{sec:motivation}


In recent years, the crucial role of data has largely been shadowed by the field's focus on architectures and training procedures.
As a result, there are no guiding principles for creating a good data set.
\textit{What are some intuitions that we get from the literature? Are they correct? What are promising future research directions?} In this paper we focus on the aspects of data quality as resulting from empirical methods for predicting generalisation, namely from Intrinsic Dimension (ID) based methods and Mixed Sample Data Augmentation (MSDA) based methods.
We show that they provide misleading insights into how one should create and manipulate data to improve model performance.

\textbf{Intrinsic Dimension:} It is believed that data lies on a low-dimensional manifold. 
Manifold's dimension should reflect the minimum number of variables required to describe the true data.
This is dependent on the task at hand.
For reconstruction, we would expect a more complex representational space than for classification. 
In this paper we focus on the latter.
While it is difficult to know the true ID, a number of estimates have been proposed~\citep[e.g.][]{granata2016accurate, duan2018bayesian, facco2017estimating, denti2021distributional}. 
Given a good model we can estimate the ID based on its representations by measuring how much its embedding space can be ``compressed''. While this is dependent on the quality of both the model and the estimator, in the paper we also provide a conceptual argument that abstracts away from these details.
It has been claimed that the lower the dimension of the manifold, the easier it is to generalise. Based on this belief, train-time generalisation estimates have been proposed. Using the TWO-NN~\citep{facco2017estimating} algorithm, \citet{ansuini2019intrinsic} estimate the global ID of last hidden layer manifolds using the train data. They then claim that the generalisation performance can be predicted based on this quantity. As such, better performance should correspond to a lower ID value. The intuition that results from this is that creating lower-dimensional data leads to better generalisation. 
Note that the  \textit{estimated} ID could be change by altering either the architecture or the data.
Since in this paper we are interested in the role and attributes of the data, we analyse the data for a fixed model.

\textbf{Mixed augmentation:} In statistical learning, training with augmented data is seen as injecting prior knowledge about the neighbourhood of the data samples.
The intuition behind augmentation caused researchers to interpret its effect through the similarity between original and augmented data distributions.
This perspective is often challenged by methods which, despite generating samples that do not appear to fall under the distribution of natural images, lead to strong learners.
This is particularly the case for MSDA, where two or more images are combined to obtain a new training sample. 
Visual examples can be found in Figure~\ref{fig:distortions}.
\citet{gontijo2020affinity} argue it is the \textit{perceived} distribution shift that needs to be minimised, while maximising the sample vicinity.
Formalising these concepts, they introduce 
augmentation ``diversity'' and ``affinity''.
Diversity is defined as the training loss when learning with artificial samples, while affinity quantifies the difference between the accuracy on original test data and augmented test data for a reference model.
The latter penalises augmentations that introduce artificial information to which the model is not invariant, implicitly assuming that training with that information is detrimental to generalisation.
In other words, it implies that preserving the data distribution when distorting data is necessary.

In this paper we show that data sets which have a higher intrinsic dimension could lead to better performance than their low-dimension counterparts. Thus, \textit{minimising data dimension is not a relevant goal when creating and refining data sets}.
Further, we construct empirical counter-examples which disprove common beliefs in the literature and highlight the importance of understanding the changes MSDAs introduce. We show that, in contrast to what is widely assumed, \textit{not preserving the data distribution can lead to learning better representations}. A direct consequence is that when dealing with limited data, the focus of the practitioners should be on understanding the changes that mixed augmentations cause, rather than choosing the augmentation that produces the smallest distribution shift. 
Correctly understanding the impact of the increasingly popular mixed-sample augmentation is essential for trusting its usage in sensitive applications where the data can be out of distribution. But most importantly, we believe this could set a new direction in capturing the relationship between data and learned representations, which could ultimately play a small role in understanding generalisation and creating better data sets.

We focus on two MSDAs, \mixup~\citep{zhang2018mixup} and \fmix~\citep{harris2020understanding}.
\mixup interpolates between two images to obtain a new sample. \fmix masks out a region of an image with the corresponding region of another image, sampling the mask from Fourier space.
We refer to models by the augmentations they were trained with and use ``basic'' for models trained without MSDA.
We do 5 runs of each experiment.  

    



\section{Should we aim to obtain a data set with minimum ID?}

\begin{table*}[t]
\begin{minipage}{0.55\linewidth}

\begin{tabularx}{\linewidth}{llll}
    \multicolumn{4}{l}{\includegraphics[width=0.99\linewidth]{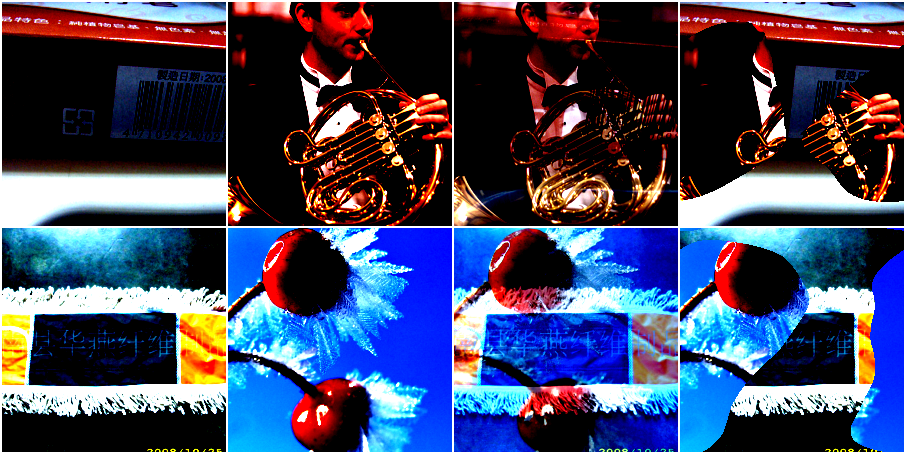} }\\
    Image 1 \hspace{1em} & Image 2 \hspace{0.8em} & \mixup & \hspace{0.8em} \fmix \\
    
\end{tabularx}
    \caption{Examples of images obtained using \mixup and \fmix augmentations with a 0.5 mixing coefficient.}
    \label{fig:distortions}
\end{minipage}
\hfil
\begin{minipage}{0.4\linewidth}
    \centering
    \caption{ ID and generalisation performance on \cifar{10} (top) and \cifar{100} (bottom). R\mixup leads to worse generalisation but lower ID.
    }\label{tab:ID}
    \begin{tabulary}{\linewidth}{lLL}
    \toprule
    & ID & Accuracy\\
    \midrule
    basic & $7.80_{\pm 17}$ & $93.04_{\pm 0.17}$ \\
    \mixup & $\bm{9.14_{\pm 0.31}}$ & $\bm{93.79_{\pm 0.18}}$ \\
    R\mixup & $7.80_{\pm 16}$ & $92.40_{\pm 0.34}$\\
    \midrule
    \midrule
    basic & $ 12.18_{\pm 1.30}$  & $71.70_{\pm0.37}$\\
    \mixup& $\bm{14.11_{\pm 1.31}}$ & $\bm{72.60_{\pm0.63}}$\\
    R\mixup  & $10.71_{\pm 0.21}$& $69.00_{\pm 0.41}$\\
    \bottomrule
    \end{tabulary}
\end{minipage}
\end{table*}

\citet{ansuini2019intrinsic} observe a correlation between data dimension and generalisation capacity. If such a correlation exists, it could be used to characterise and improve data sets.
The method they use to compute ID requires training a model, making it impractical for the purpose of data creation and refinement. 
However, if intrinsic dimension can indeed be used as a generalisation capacity predictor, the effort of the community could be steered towards building more efficient estimators. 
But is lower ID the driving factor of stronger learners or is this correlation coincidental? 
In this section we show that higher ID representations can lead to better generalisation performance, thus disproving the above correlation.
We train the same model architecture on data with different ID and compute the estimate of representation dimension.
Following \citet{ansuini2019intrinsic}, we use the TWO-NN estimator introduced by \citet{facco2017estimating}, where the ratio of the distances to the closest two neighbours of each point are used to approximate manifold dimension.
We then argue that even if the representation ID estimator was not entirely reflective of true data dimension, minimising data ID does not imply better generalisation and is not a relevant objective when creating data sets.

To obtain data with different properties in a controlled manner, we make use of augmentation.
In addition to the two MSDAs introduced above, we also create a data set with a variation of \mixup equivalent to an objective reformulation \cite{huszar2017, harris2020understanding} which we label R\mixup. 
This consists of creating a new sample by interpolating two images but unlike \mixup, the new sample is assigned the label of only one of the source images. 
Ignoring one of the targets when mixing inputs is expected to create a data set where the instances can be represented in a more compressed manifold, decreasing separability at the same time.
Table~\ref{tab:ID} shows the results we obtain for the VGG16~\citep{Simonyan15} network on the \cifar{10/100}~\citep{krizhevsky2009learning} data sets. \mixup has the highest test accuracy, while having a significantly higher ID compared to the R\mixup model. 
\textit{This directly contradicts the idea that a minimum ID data set is necessarily better}. 

One question that is immediately raised is if our conclusion would still hold given a more accurate method of capturing manifold dimension.
We argue that even with further estimator refinements, this hypothesis lacks a strong basis and it is unlikely to hold in practice. 
To see this more clearly, we can think of a binary image classification, where a data collection artefact is present for one of the classes such as a specific small group of identifying pixels. In this case, a learner that classifies entirely based on this spurious rule would achieve very high compression with no real generalisation abilities.
Thus, we should not seek to minimise the intrinsic dimension when processing data.

So what should one seek when manipulating data? A line of work that tries to address this is augmentation analysis. There is no unifying framework for understanding augmentation. Despite the lack of consensus in the field, there is one undisputed belief that \textit{the smaller the distribution gap between original and manipulated data, the better the model generalises}. In the following section we focus on this belief which impacts not only augmentation, but data manipulation as a whole.


\section{Is the magnitude of the distribution shift important when manipulating data?} \label{sec:MSDA}



\begin{table}[t]
    \centering
    \caption{Augmentation comparison on \cifar{10}. We consider two variants when calculating diversity. One is the cross-entropy loss using the label of the majority class (Diversity), as for mixing in \cite{inoue2018data}. The alternative, MixDiversity, takes a linear combination of the two cross-entropy losses.} \label{tab:affinity}
    \begin{tabulary}{\linewidth}{llll}
    \toprule
    & Affinity & Diversity & MixDiversity\\
    \midrule
    \mixup  & $\mathbf{-12.58_{\pm 0.14}}$ & $\mathbf{0.41_{\pm 0.01}}$ & $\mathbf{0.84_{\pm 0.00}}$\\
    \fmix & $-25.55_{\pm 0.26}$ & $0.34_{\pm 0.01}$ & $0.65_{\pm 0.00}$\\
    \bottomrule
    \end{tabulary}
\end{table}

Traditionally it was believed that a good augmentation should have minimal distribution shift.
Most recently, it has been argued that it is the degree of the \textit{perceived} shift that determines augmentation quality \cite{gontijo2020affinity}.
We show that \textit{the magnitude of the distribution shift does not determine augmentation quality}. 
We start with the perceptual gap of training with MSDA, as proposed in \citet{gontijo2020affinity}. Reiterating, this is given by the difference between the performance of the baseline model when presented with original test data and augmented test data and is termed ``affinity''. 
Subsequently, we address the gap in the wider sense, as is often sought in prior art.
We first argue that high affinity and high diversity are not necessarily desirable.
Indeed, on \cifar{10}, we find \fmix, a better performing augmentation, to have both lower affinity and lower diversity than \mixup (Table~\ref{tab:affinity}). 
For diversity, we compute the cross-entropy loss where the label is taken to be that of the majority class.
Similar results are obtained with the \mixup loss, where a weighted average of the true labels is taken. 

While intuitively for a high level of affinity, high diversity could correspond to better methods, the converse does not hold.
We argue this is because affinity is rather an analysis of the learnt representations of the reference model and cannot give an insight into the quality of the augmentation or its effect on learning. 
As such, an augmentation will have a lower affinity if it introduces artefacts that could otherwise lead to learning better representations when used in the training process.
We believe this issue extends to other approaches that aim to motivate the success of MSDA through reduced distribution shift.
Henceforth, we focus on bringing further supporting evidence that the importance lies in the invariance introduced by the shift and its interaction with the given problem rather than its magnitude.


    

\subsection{If it is not the magnitude that matters, is it the direction?} \label{sec:bias_matters}
We use empirical evidence to argue against previous assumptions behind the success of MSDA and propose the study of introduced bias as a more informative research direction.
We use the term ``bias'' to refer to a drift in the learnt representations introduced by the change in the training procedure.
A fundamental difference to classical training is that the samples are no longer independent when augmenting.
Mixed-sampling takes this even further.
An immediate question is, does the added correlation lead to more meaningful representations? 
It is claimed that the strength of \mixup lies in causing the model to behave linearly between two images~\citep{zhang2018mixup} or in pushing the examples towards their mean~\citep{carratino2020mixup}. 
Both of these claims rely on the combined images to be generated from the same distribution. 
Performing inter-dataset augmentation we show that this is not necessary for a successful augmentation. The same experiment further shows that by distorting the data distribution by the same magnitude we can obtain two different results depending on the direction of the introduced bias. 

We once again use the reformulated objective setting, where two images are mixed without mixing targets as well.
This allows us to apply MSDA between data sets. 
Thus, for training a model on a data set, we use an additional one whose targets will be ignored.
As an example, a model that is learning to predict \cifar{10} images will be trained on a combination of \cifar{10} and \cifar{100} images, with the target of the former.
This scenario breaks the added correlation between training examples. 

Table~\ref{tab:other} contains the results of this experiment, showing that an accuracy similar to or better than that of regular MSDA can be obtained by performing inter-dataset MSDA. 
This invalidates the argument that the power of \mixup resides in causing the model to act linearly between samples.
Another observation is that for \fmix and \mixup, introducing elements from \cifar{100} when training models on the \cifar{10} problem does not harm the learning process.
The reciprocal, however, does not hold.
Hence, the ``distribution shift'' is more intimately linked to the problem at hand and aiming to characterise an augmentation based on the distance from the original distribution is a limiting approach.
This experiment shows that shifting two distributions by the same amount can have different effects on the model performance.
Thus, \textit{the specifics of the bias introduced could be more important than its magnitude}.
While some level of data similarity has to be preserved when performing MSDA, it is far from being the objective of such data-distorting approaches which should be rather seen as forms of regularisation (see Appendix~\ref{sec:complexity} for experiments and discussion on the increase in data complexity added by MSDA).


\begin{table*}[t]
    \centering
    \caption{Accuracy on \cifar{10} (left) and \cifar{100} (right) upon mixing with samples from a different data set. The baseline is the accuracy when training with a single data set. \cifar{110} is used to refer to mixing with \cifar{100} when training on the \cifar{10} problem and vice-versa. 
    }\label{tab:other}
    \begin{tabulary}{\linewidth}{lLLLLL}
    \toprule
    & \mixup & \fmix & & \mixup & \fmix\\
    \midrule
    baseline  & $94.18_{\pm0.34}$ & $94.36_{\pm 0.28}$ & &$74.68_{\pm 0.37}$  & $75.75_{\pm 0.31}$\\ 
    \cifar{110} & $94.70_{\pm 0.27}$ & $94.80_{\pm 0.32}$ & & $72.36_{\pm1.04}$ & $74.80_{\pm 0.55}$ \\ 
    Fashion & $92.28_{\pm0.28}$ & $95.03_{\pm 0.10}$ & & $66.40_{\pm 1.86}$ & $74.46_{\pm0.57}$\\
    \bottomrule
    \end{tabulary}
\end{table*}

In this paper we demonstrate that the shift in learnt representations can lead to better models and simply quantifying the distribution shift can be misleading.
An open question remains: How can we better capture the bias that is introduced and measure its quality? We believe understanding how a relatively small change in the data distribution impacts learnt representations could lead the way to characterising the relationship between data and model generalisation. 


\section{Conclusions}

Starting from generalisation studies, we empirically disprove the hypothesis that lower data dimension is necessarily associated with better performance on unseen data. 
We then show that the purpose of data manipulation is not to leave the distribution unchanged, but to modify it in a principled and constructive manner. 
The focus of the community must be on analysing the introduced bias rather than its elimination.
Correctly interpreting this bias is important not only for making the models trustable but also for injecting more informed prior knowledge in future applications.
Beyond their practical benefits, we believe MSDAs have the potential to help characterise the interplay between data and learnt representations.

\bibliography{References}







\appendix
\section{MSDA increases data complexity} \label{sec:complexity}
We believe that MSDA training could help bypass some of the simplicity bias.
The simplicity bias refers to the tendency of deep models to find simple representations and has been used to justify the success of deep models~\citep{nakkiran2019sgd, valle-perez2018deep}.
Recent research shows that this propensity causes models to ignore complex features that explain the data well in favour of elementary features, even when they lead to worse performance \citep{NEURIPS2020_6cfe0e61, hermann2020shapes}.

Although it could seem natural that MSDAs increase the complexity of the problem, we design an experiment to support this claim.
Similarly to~\citet{NEURIPS2020_6cfe0e61}, we combine \cifar{10} and MNIST~\citep{lecun-mnisthandwrittendigit-2010} samples.
Since they have the same number of classes, we can easily associate each class of one data set with a corresponding one from the other.
Thus, we stack a padded image from the $k$th class of MNIST on top of a sample from the $k$th class of \cifar{10}, such that a $3 \times 64 \times 32$ image is obtained.
We then randomly combine the test images and separately compute the accuracy with respect to the targets of each data set. 

The predictions with respect to the \cifar{10} labels are no better than random ($10.04_{\pm 0.11}$), while the accuracy with respect to the MNIST images remains high ($99.57_{\pm 0.72}$).
Thus, models trained on this combination are mostly relying on MNIST images to make predictions. 
Similar behaviours have previously been associated with simplicity bias.
Subsequently, when training, we perform \fmix only on MNIST images and observe that this is enough to reverse the results. 
Evaluating against the \cifar{10} label gives an accuracy of $86.60_{\pm 0.34}$, while testing against the MNIST label only gives $11.61_{\pm 0.30}$. 
We find that this also holds true for the other MSDAs.
Thus, performing these distortions on the simpler data set increases its complexity to the point where it surpasses that of \cifar{10}.

\appendix
\appendix
\section{Experimental details} \label{sup:exp_details}
Throughout the paper, we use PreAct-ResNet18~\citep{he2016identity} models, trained for 200 epochs with a batch size of 128.
For the MSDA parameters we use the same values as~\citet{harris2020understanding}.
All models are augmented with random crop and horizontal flip and are averaged across 5 runs.
We optimise using SGD with 0.9 momentum, learning rate of 0.1 up until epoch 100 and 0.001 for the rest of the training. 
This is due to an incompatibility with newer versions of the PyTorch library of the official implementation of~\citet{harris2020understanding}, which we use as a starting point for model training. 
However, the difference in learning rate schedule between our work and prior art does not affect our findings since we are not introducing a new method to be applied at training time.
In our case, it is sufficient to show that the bias exists in at least one configuration. 
For the analysis we also used adapted code from~\citep{carlucci2019domain} for patch-shuffling.
The models were trained on either one of the following: Titan X Pascal, GeForce GTX 1080ti or Tesla V100. 
For the analyses, a GeForce GTX 1050 was also used. The average training time was less than two hours, with the exception of model trained on Tiny-\imagenet, which took around 10 hours to run.

\subsection*{Training models}
The code for model training is largely based on the open-source official implementation of \fmix, which also includes those of \mixup, \cutout, and \cutmix.
For the experiment where we use the reformulated objective to combine data sets, instead of mixing with a permutation of the batch, as it is done in the original implementation of the mixed-augmentations, we now draw a batch form the desired data set. 
To ensure a fair comparison, for the basic we also perform inter-batch mixing.

\subsection*{Evaluating robustness}

For the \cutocc measurement, we modify open-source code to restrict the occluding patch to lie withing the the margins of the image to be occluded. This is to ensure that the mixing factor $\lambda$ matches the true proportion of the occlusion.
For \iocc, the implementation of Grad-CAM is again adapted from publicly available code.
With both methods, we evaluate 5 instances of the same model and average over the results obtained.

The added computation time of \iocc over the regular \cutocc for a fixed occlusion fraction is that of performing Grad-CAM on train and test data, as well as evaluating on the latter.
With a batch size of 128, this takes under half an hour. 

\section{Analysis of wrong predictions} 

\subsection{Alternative index} \label{sup:alt_idx}
Table~\ref{tab:sup-sh-tx-bias} the worst-case DI index where we replace $i_{c_{max}}$ in Equation 1 by the maximum increase across the runs.
As per the original formulation, we note that the masking methods lead to models which are less sensitive to the artefacts resulted after patch-shuffling.


\subsection{Varying the grid size} \label{sup:dif_grid_size}

Table~\ref{tab:alt_grid_size} gives the results obtained when varying the number of image tiles to be randomly rearranged. We observe that data interference appears for different grid sizes.

\begin{table}
    \begin{minipage}{0.61\linewidth}
        \centering
    \caption{Alternative DI index for PreAct-ResNet18 on grid-shuffled images for four different types of models. Again, a bias can be noted for all considered data sets.} \label{tab:sup-sh-tx-bias}
    \begin{tabulary}{\linewidth}{lLLLL}
    \toprule
            & basic  & \mixup & \fmix   & \cutmix     \\
     \midrule
    \cifar{10} & $3.52_{\pm 0.56}$ &  $3.31_{\pm 0.82}$ & $0.76_{\pm 0.16}$  & $0.43_{\pm 0.13} $ \\
    \cifar{100} & $1.40_{\pm 0.38}$ &$1.09_{\pm 0.29}$ &
$0.38_{\pm 0.21}$ &$0.16_{\pm 0.08}$ \\
    FashionMNIST & $1.56_{\pm 0.39}$ & $3.57_{\pm 1.35}$ & 
$1.65_{\pm 0.35}$ & $0.82_{\pm 0.13}$ \\
    Tiny \imagenet & $3.01_{\pm 0.48}$ & $2.24_{\pm 0.30}$ & $2.34_{\pm 1.86}$ & $11.45_{\pm 10.54}$  \\
    \imagenet & $0.82$ & $1.49$ & $0.58$ & $-$\\
    
     \bottomrule
    \end{tabulary}
    \end{minipage}
    \hfil
    \begin{minipage}{0.34\linewidth}
        \centering
    \caption{Shape and texture accuracy of BagNet9 models on the GST data set.}
    \label{sup_tab:tiny_bag}
    \begin{tabulary}{\linewidth}{lLL}
    \toprule
                & Shape             & Texture \\
    \midrule
    basic    & $11.29_{\pm 0.15}$ & $18.90_{\pm0.66}$ \\
    \mixup      & $11.04_{\pm 0.29}$ & $12.56_{\pm 1.26}$\\ 
    \fmix       & $11.06_{\pm 0.48}$ & $17.47_{\pm1.74}$\\
    \cutmix     & $10.76_{\pm 0.27}$ & $20.28_{\pm 0.88}$\\
    \bottomrule
    \end{tabulary}
    \end{minipage}
\end{table}

\begin{table}
        \centering
    \caption{DI index for alternative grid sizes.} \label{tab:alt_grid_size}
    \begin{tabulary}{\linewidth}{lLLLLL}
    \toprule
            & & basic  & \mixup & \fmix   & \cutmix  \\
    \midrule
    \multirow{2}{*}{\cifar{10}} & $2 \times 2$ & $0.61_{\pm 0.24}$ & $0.56_{\pm 0.33}$ & $0.19_{\pm 0.14}$ & $0.12_{\pm 0.06}$ \\
                & $8 \times 8$ & $6.41_{\pm 0.55}$ & $6.95_{\pm 1.96}$ & $2.75_{\pm 1.46}$ & $1.41_{\pm 1.15}$ \\
    \midrule
    \multirow{2}{*}{\cifar{100}} & $2 \times 2$ & $1.03_{\pm 0.29}$ & $0.46_{\pm 0.14}$ & $0.21_{\pm 0.14}$ & $0.12_{\pm 0.07}$ \\
                & $8 \times 8$ & $9.16_{\pm 6.15}$ & $3.10_{\pm 4.59}$ & $1.62_{\pm 0.89}$ & $0.65_{\pm 0.50}$ \\
     \midrule

    \multirow{2}{*}{Tiny~\imagenet} & $8 \times 8$ & $5.76_{\pm 6.61}$ & $5.73_{\pm 3.82}$ & $2.49_{\pm 1.38}$ & $0.60_{\pm 0.69}$ \\
    & $16 \times 16$ &  $44.01_{\pm 36.47}$& $14.06_{\pm 14.63}$ &
$11.94_{\pm 17.79}$& $1.86_{\pm 1.98}$ \\
    \midrule
    \multirow{2}{*}{\imagenet} & $4 \times 4$ & $0.82$ & $1.49$ & $0.58$ & $-$ \\
    & $64 \times 64$ & $4.89$ & $41.16$ & $12.77$& $-$\\
    \bottomrule
    \end{tabulary}
\end{table}

\subsection{Patch-shuffling} \label{sup:wrongPred_shape}

\begin{figure*}
    \centering
    \includegraphics[width=0.8\linewidth]{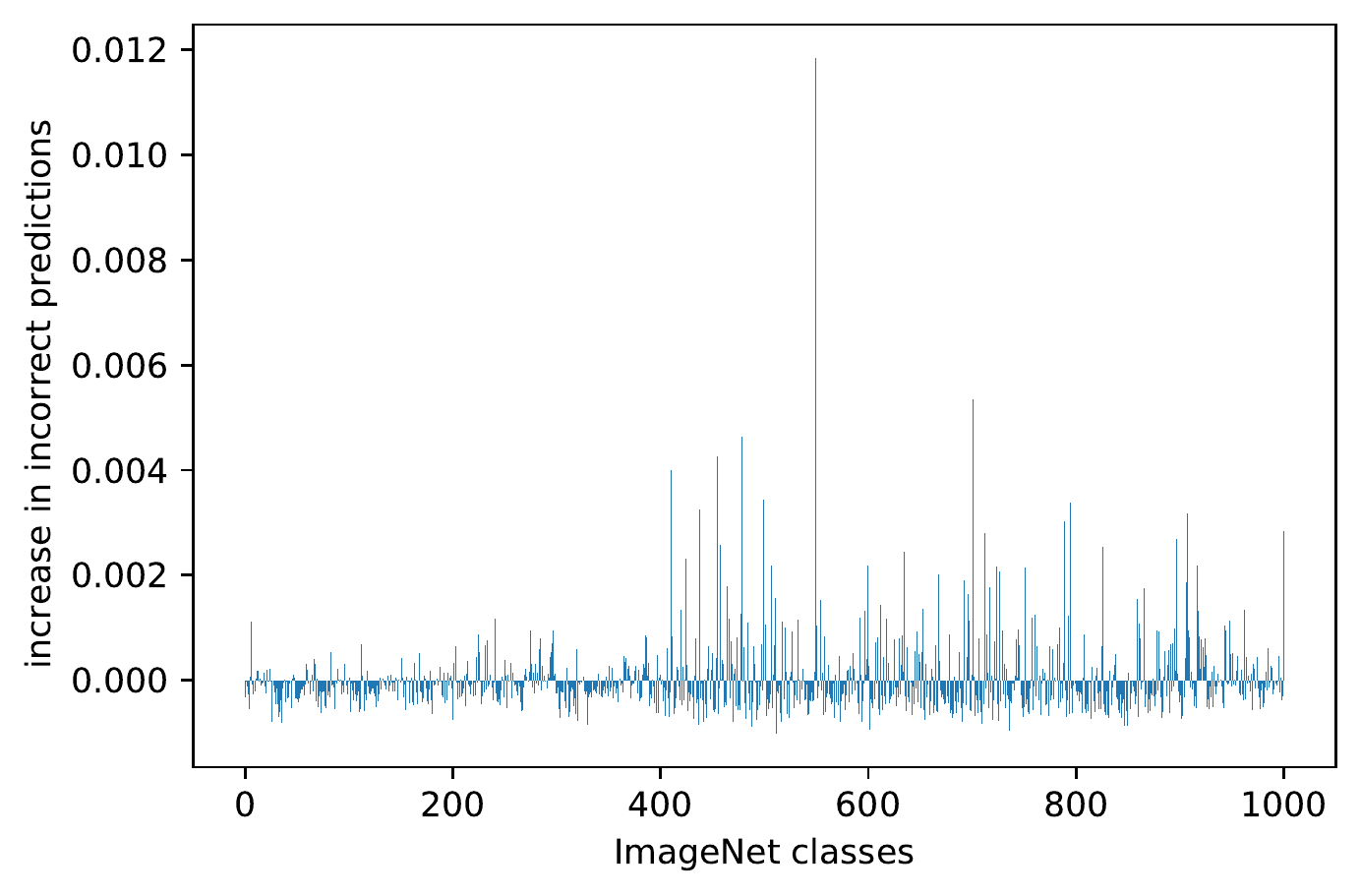}
    \caption{Difference between the number of times a class was wrongly predicted when presented with regular \imagenet samples and patch-shuffled data.} 
    \label{fig:imnet}
\end{figure*}

We look at the classes which have the highest increase in incorrect predictions and note that their shapes are characterised by strong horizontal and vertical edges.
For example, on \cifar{100}, varying the grid size between $2 \times 2$, $4 \times 4$ and $8 \times 8$ gives "Lamp", "Bus" and "Table" as dominant $c_{max}$ classes, while the model trained on Fashion MNIST with the standard procedure tends to predict grid-shuffled images as "Bag".
Figure~\ref{fig:imnet} shows that on \imagenet, the basic model tends to wrongly identify the patch-shuffled images as belonging to class "Envelope".

\subsection{\cutocc} \label{sup:more_cutocc_DI}
\begin{figure*}
    \begin{subfigure}[t]{0.49\linewidth}
        \centering
        \includegraphics[width=\linewidth]{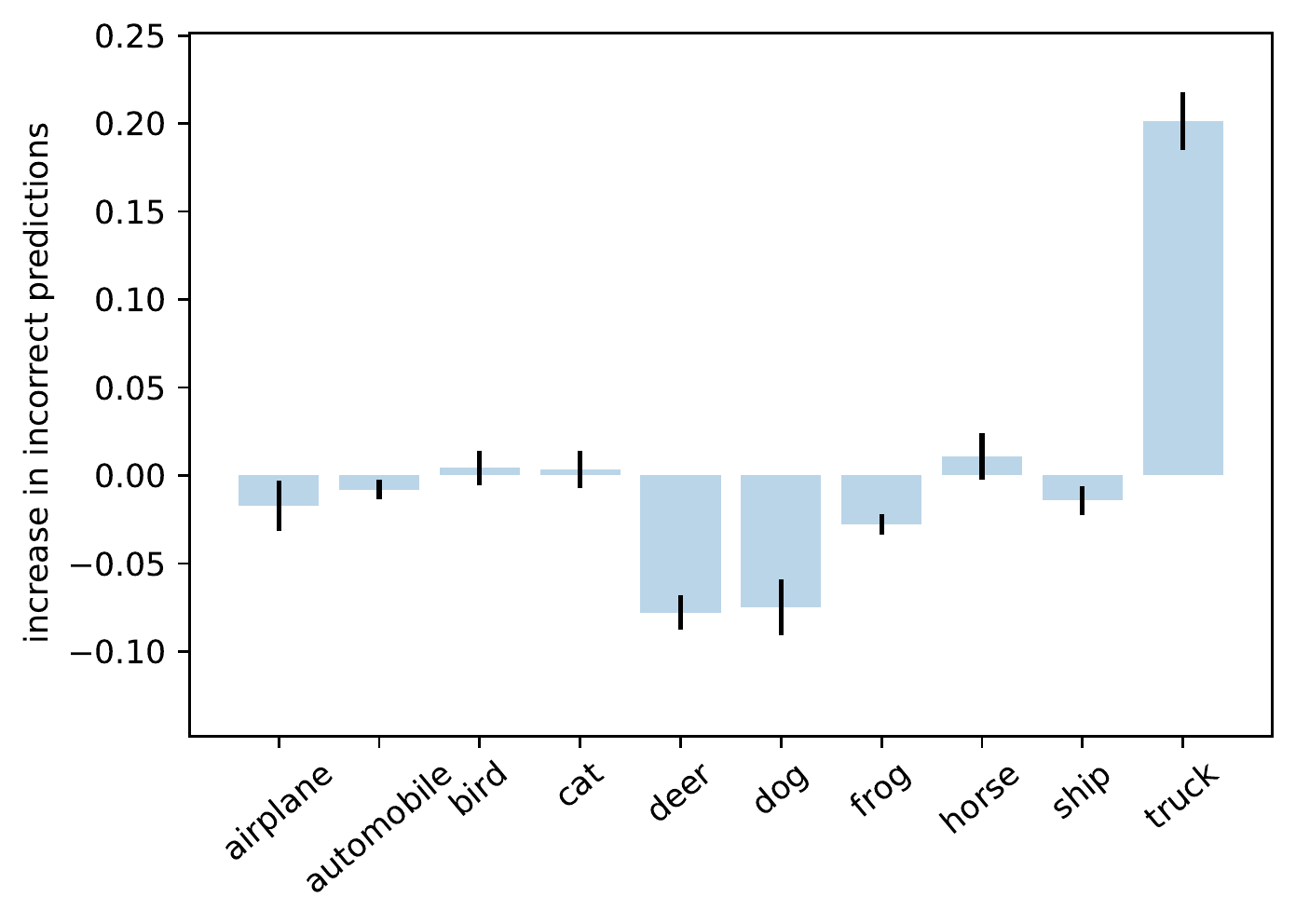}
        \caption{basic}
    \end{subfigure}
    \hfill 
    \begin{subfigure}[t]{0.49\linewidth} 
        \centering
        \includegraphics[width=\linewidth]{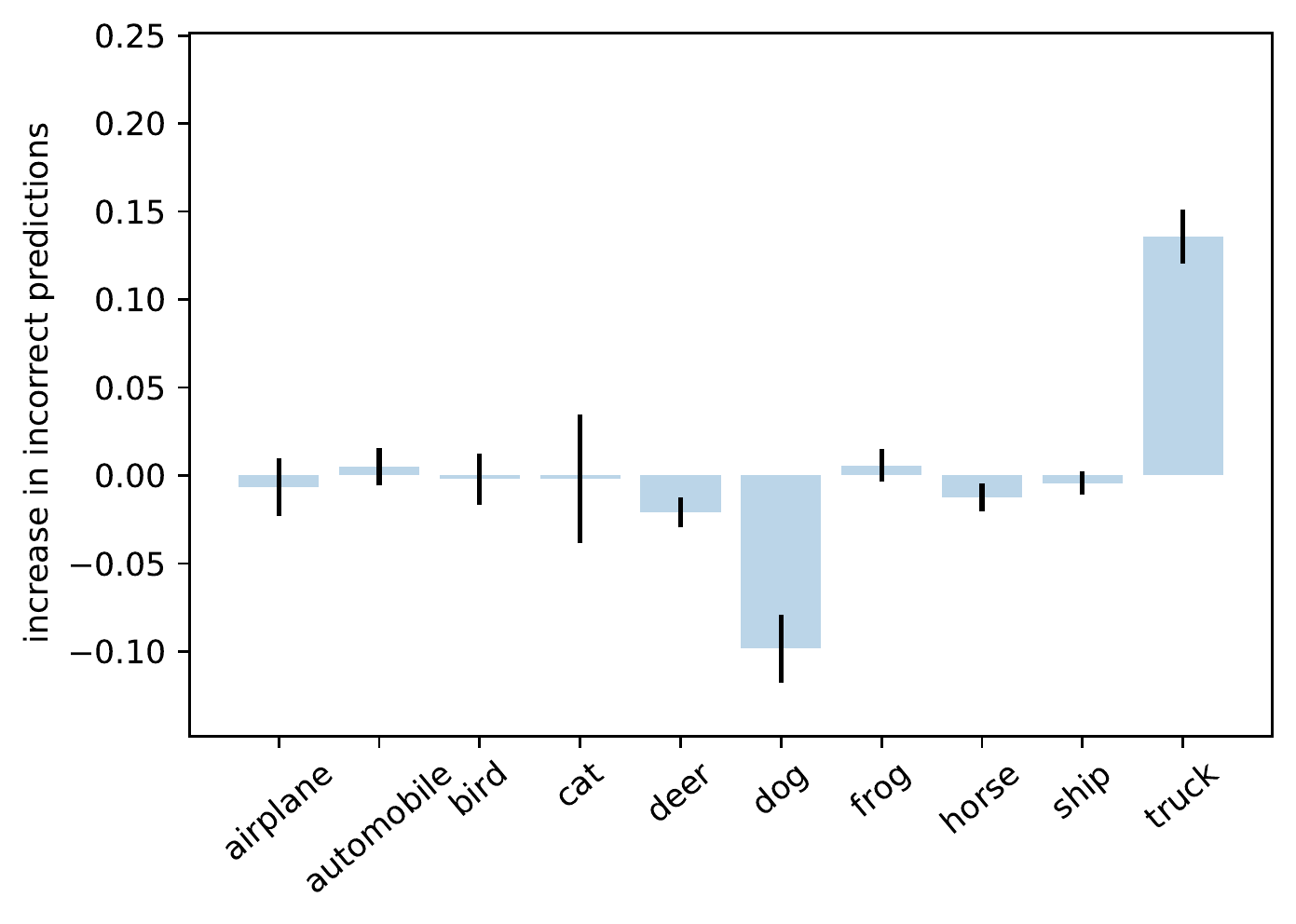}
        \caption{\mixup} 
    \end{subfigure}
    \newline
    \begin{subfigure}[t]{0.49\linewidth} 
        \centering
        \includegraphics[width=\linewidth]{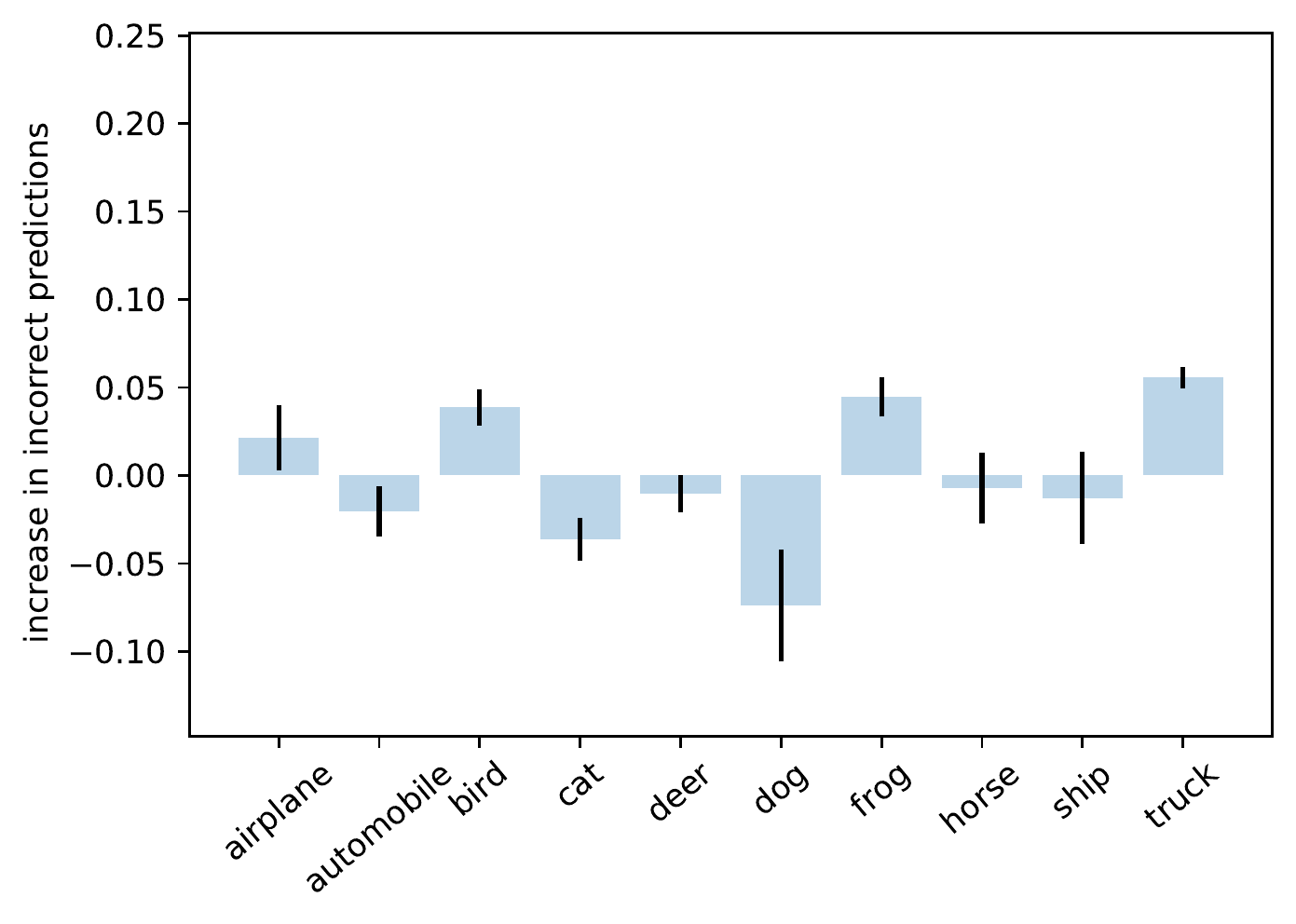}
        \caption{\cutmix} 
    \end{subfigure}
    \hfill 
    \begin{subfigure}[t]{0.49\linewidth} 
        \centering
        \includegraphics[width=\linewidth]{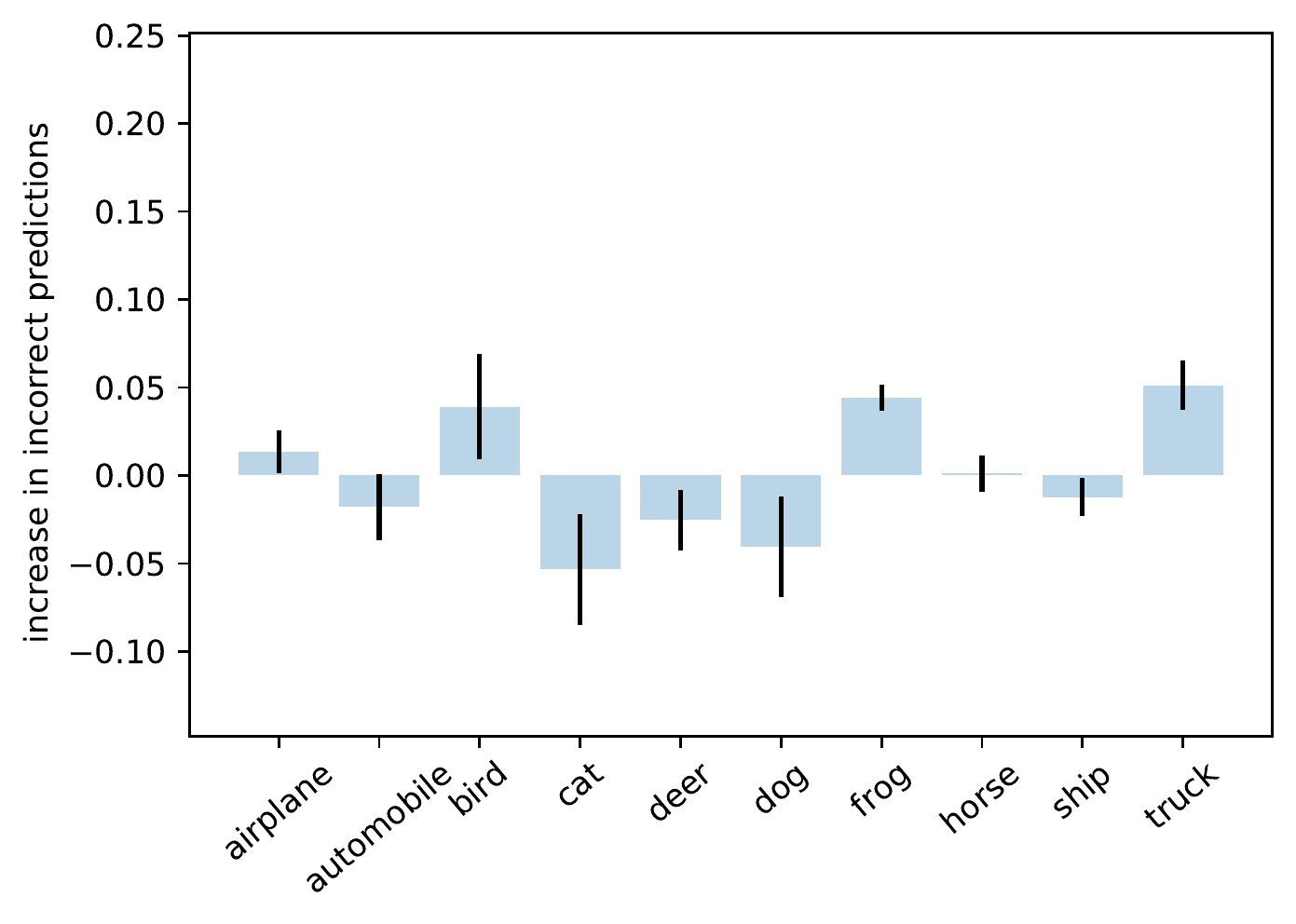}
        \caption{\fmix} 
    \end{subfigure}
    \caption{Difference between wrongly predicted classes when testing on original data versus \cutmix images. The evaluated models from left to right, top to bottom are trained on \cifar{10} with: no mixed-data augmentation (basic), \mixup, \cutmix, and \fmix.}
    \label{fig:wrong_pred_cutmix_c10}
\end{figure*}

In this section we experiment with alternative masking methods when computing \cutocc. We note that the bias exists when occluding with patches taken from images belonging to different data sets (Table~\ref{tab:cutmixing}). 
Figure~\ref{fig:wrong_pred_cutmix_c10} gives a visual account of the results obtained for \cifar{10} when mix-patching.
Note that for Fashion~MNIST we use MNIST, for Tiny~\imagenet we use \imagenet, while for \cifar{10} we mix with \cifar{100} and vice versa. Since \imagenet images are significantly larger than those of the other data sets, mixing would imply padding large areas, which would give results very similar to uniform patching.
We also experiment with VGG models, where on \cifar{10} the basic has a DI index of $0.80_{\pm 0.40} $ compared to  $0.18_{\pm 0.11} $ of \mixup.
We then use masks sampled from Fourier space (Table~\ref{tab:fout}) and note that even for these irregularly shaped distortions, we can identify a gap in most cases. The only exception is in the case of Fashion MNIST. It must be stressed that although all the models we experimented with presented Data Interference for this problem, this does not exclude the possibility of constructing a different model that is insensitive to this distortion. 
For example, we identify a gap for this problem when mix-masking (DI index of $4.09_{\pm 1.74}$ for the basic model as opposed to $1.87_{\pm 0.27}$ for a model trained on images that were masked out using \fmix-like masks).
Thus, when occluding with a particular shape we implicitly disfavour models in which learnt representations are related to the features introduced by that shape.

\begin{table}
        \centering
    \caption{DI index for occluding with images from another data set.} \label{tab:cutmixing}
    \begin{tabulary}{\linewidth}{lLLLL}
    \toprule
            & basic  & \mixup & \fmix   & \cutmix  \\
    \midrule
    \cifar{10} & $0.18_{\pm 0.05}$ &$0.39_{\pm 0.15}$ &$0.12_{\pm 0.11}$ &$0.08_{\pm 0.06}$\\
    \cifar{100} & $0.48_{\pm 0.09}$& $0.61_{\pm 0.27}$& $0.90_{\pm 0.15}$& $1.25_{\pm 0.25}$\\
    Fashion MNIST & $3.40_{\pm 0.29}$ & $3.06_{\pm 1.07}$ & $1.81_{\pm 0.55}$ & $2.61_{\pm 0.80}$\\
    Tiny~\imagenet& $0.25_{\pm 0.12}$& $0.17_{\pm 0.04}$& $0.06_{\pm 0.03}$& $0.12_{\pm 0.04}$\\
    \bottomrule
    \end{tabulary}
\end{table}

\begin{table}
        \centering
    \caption{DI index for patching using masks sampled from Fourier space.} \label{tab:fout}
    \begin{tabulary}{\linewidth}{lLLLL}
    \toprule
            & basic  & \mixup & \fmix   & \cutmix  \\
    \midrule
    \cifar{10} & $2.08_{\pm 1.13}$& $1.79_{\pm 1.09}$& $1.32_{\pm 0.99}$& $4.21_{\pm 1.23}$\\
    \cifar{100} & $4.06_{\pm01.47}$ & $3.11_{\pm02.29}$ & $9.90_{\pm14.32}$ & $2.89_{\pm05.36}$\\
    Fashion MNIST & $49.55_{\pm 20.45}$& $40.69_{\pm 21.63}$& $27.87_{\pm 17.57}$& $61.04_{\pm 17.92}$\\
    Tiny~\imagenet & $4.37_{\pm 0.85}$ &$6.95_{\pm 1.84}$ &$3.60_{\pm 1.73}$ &$5.92_{\pm 4.38}$\\
    \imagenet & $3.27$ & $2.24$ & $6.08$ & $-$\\
    \bottomrule
    \end{tabulary}
\end{table}

\begin{figure*}
    \begin{subfigure}[t]{0.49\linewidth}
        \centering
        \includegraphics[width=\linewidth]{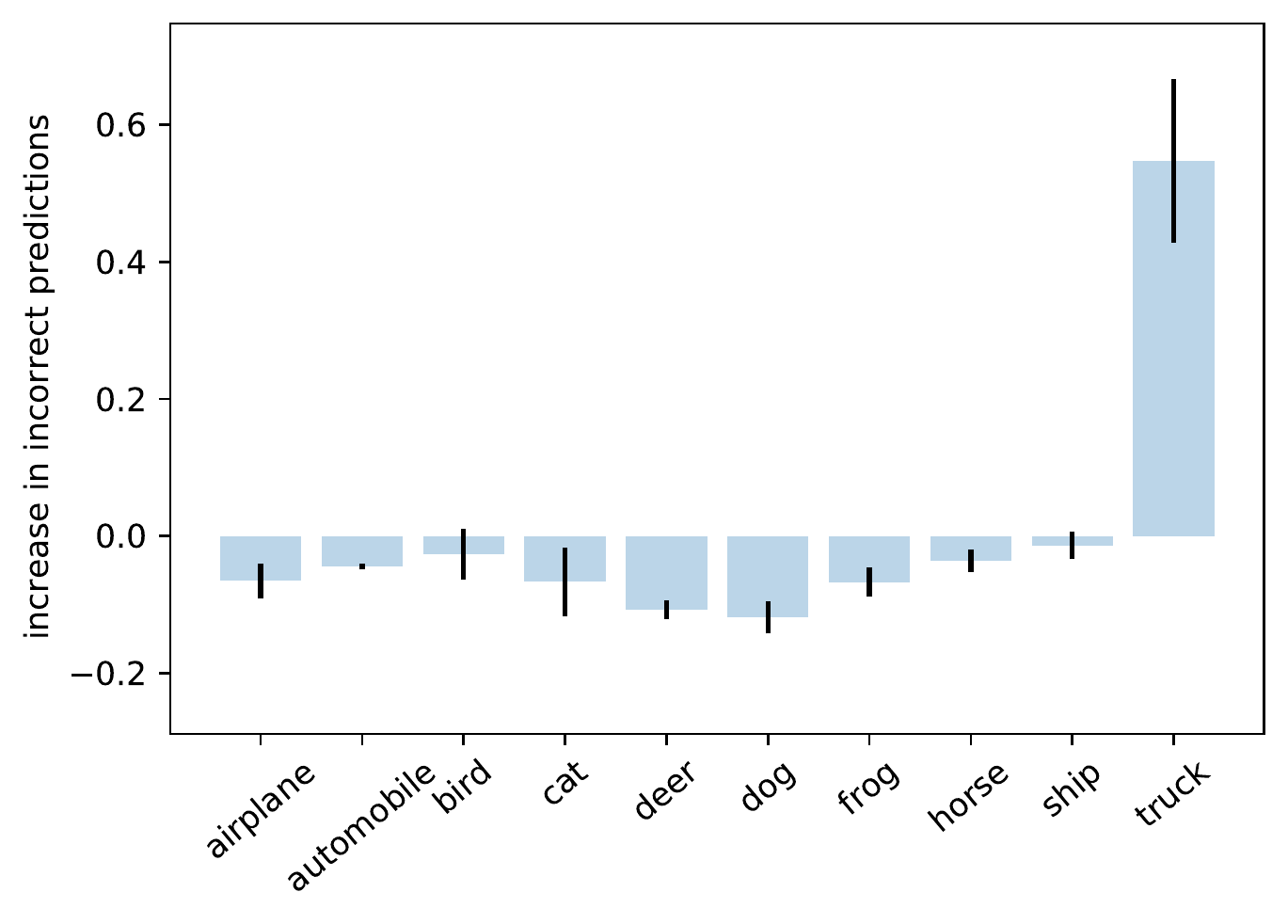}
        \caption{basic}
    \end{subfigure}
    \hfill 
    \begin{subfigure}[t]{0.49\linewidth} 
        \centering
        \includegraphics[width=\linewidth]{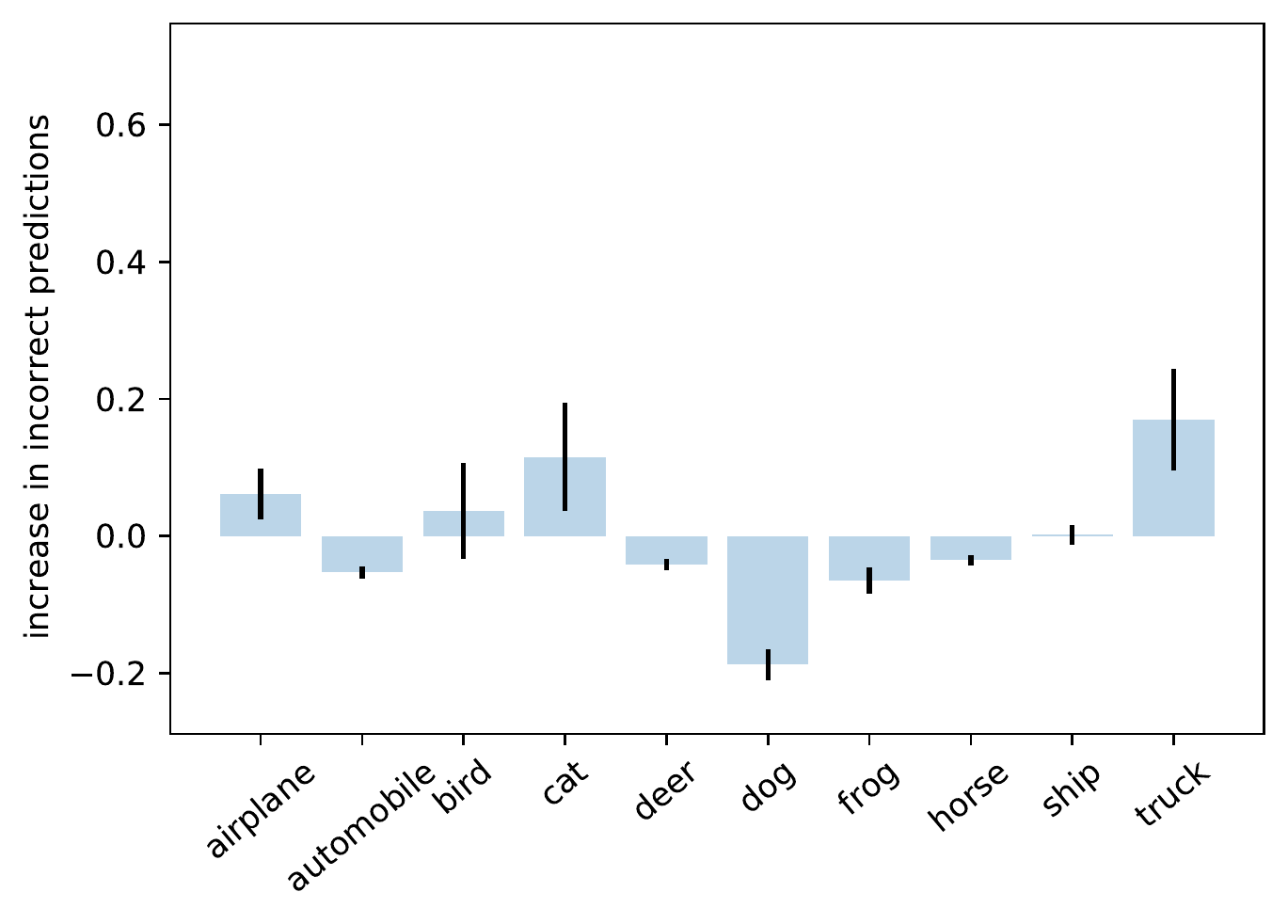}
        \caption{\mixup} 
    \end{subfigure}
    \newline
    \begin{subfigure}[t]{0.49\linewidth} 
        \centering
        \includegraphics[width=\linewidth]{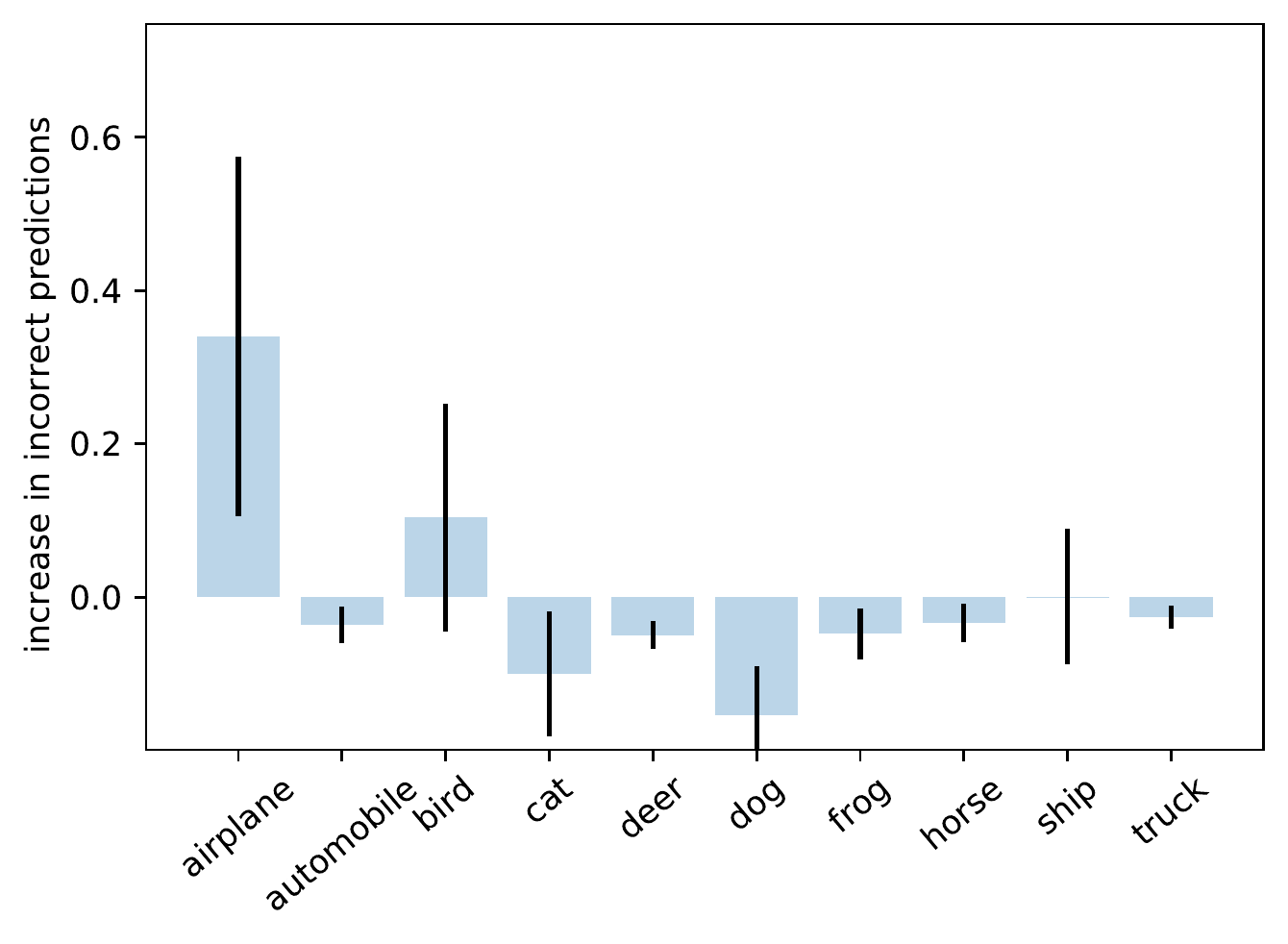}
        \caption{\cutmix} 
    \end{subfigure}
    \hfill 
    \begin{subfigure}[t]{0.49\linewidth} 
        \centering
        \includegraphics[width=\linewidth]{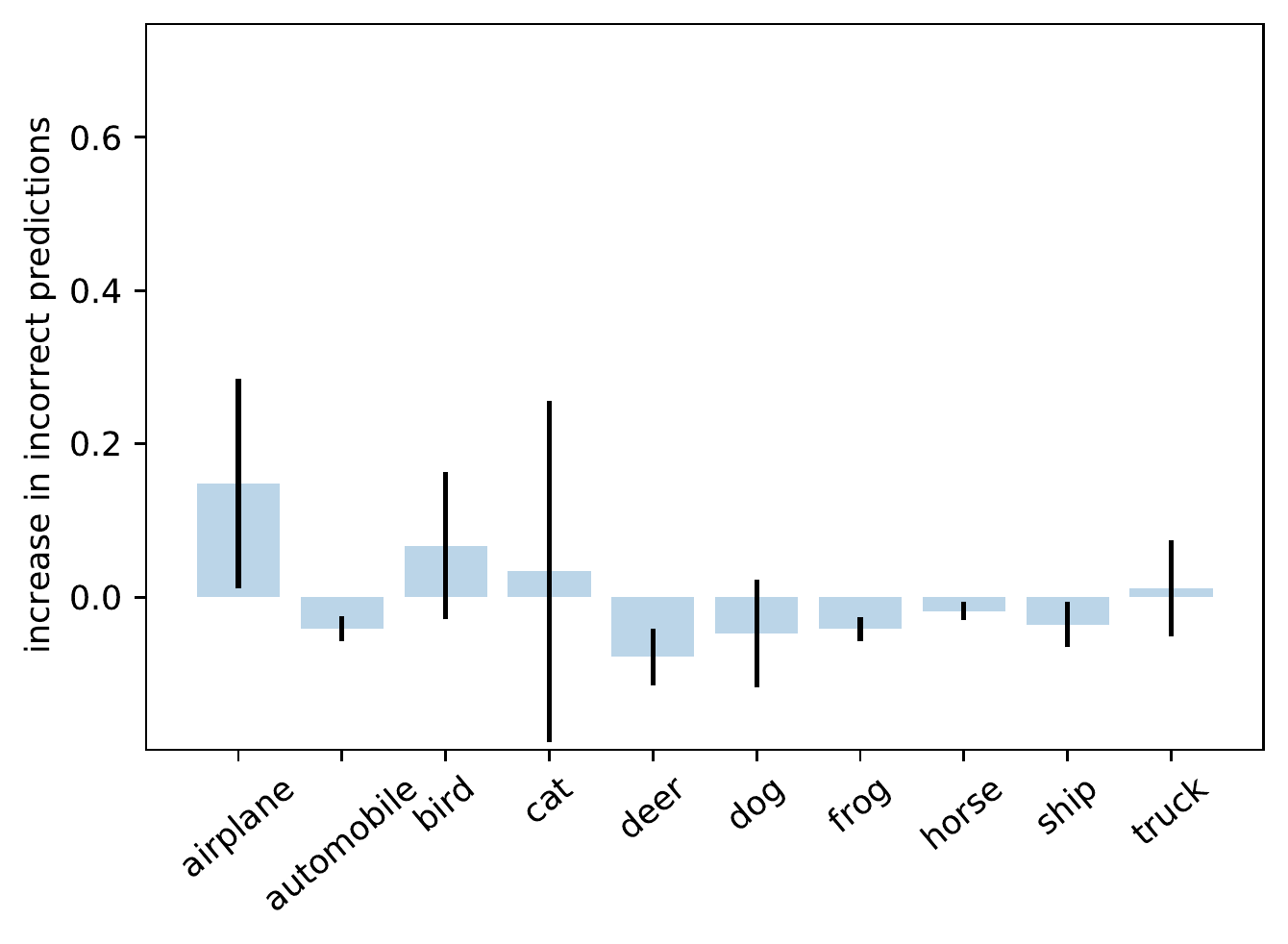}
        \caption{\fmix} 
    \end{subfigure}
    \caption{Difference between wrongly predicted classes when testing on original data versus \cutout{} images. The evaluated models from left to right, top to bottom are trained on \cifar{10} with: no mixed-data augmentation (basic), \mixup, \cutmix, and \fmix.}
    \label{fig:wrong_pred_cutout_c10}
\end{figure*}

\begin{figure*} 
    \begin{subfigure}[t]{0.49\linewidth}
        \centering
        \includegraphics[width=\linewidth]{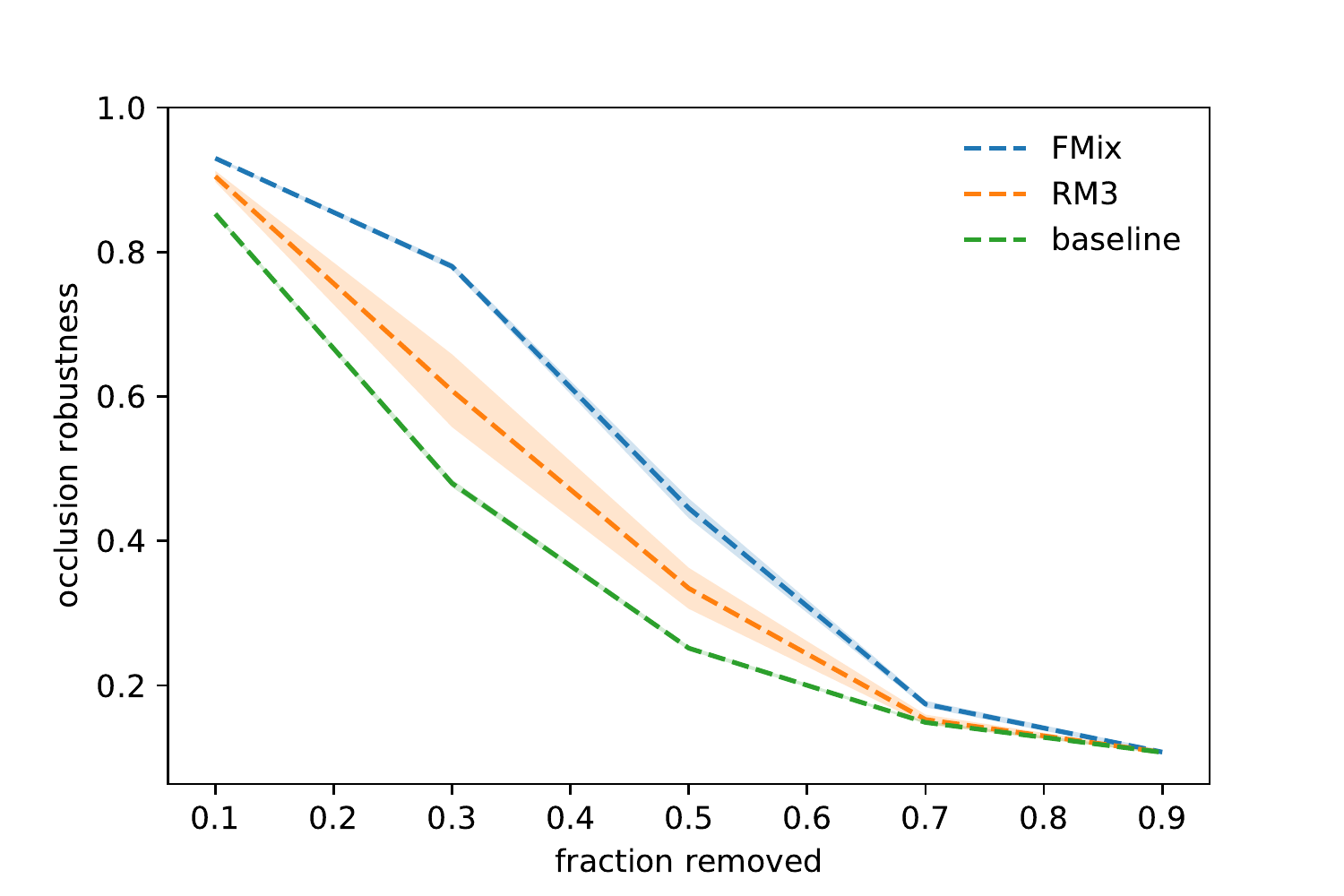}
    \end{subfigure}
    \hfill 
    \begin{subfigure}[t]{0.49\linewidth}
        \centering
        \includegraphics[width=\linewidth]{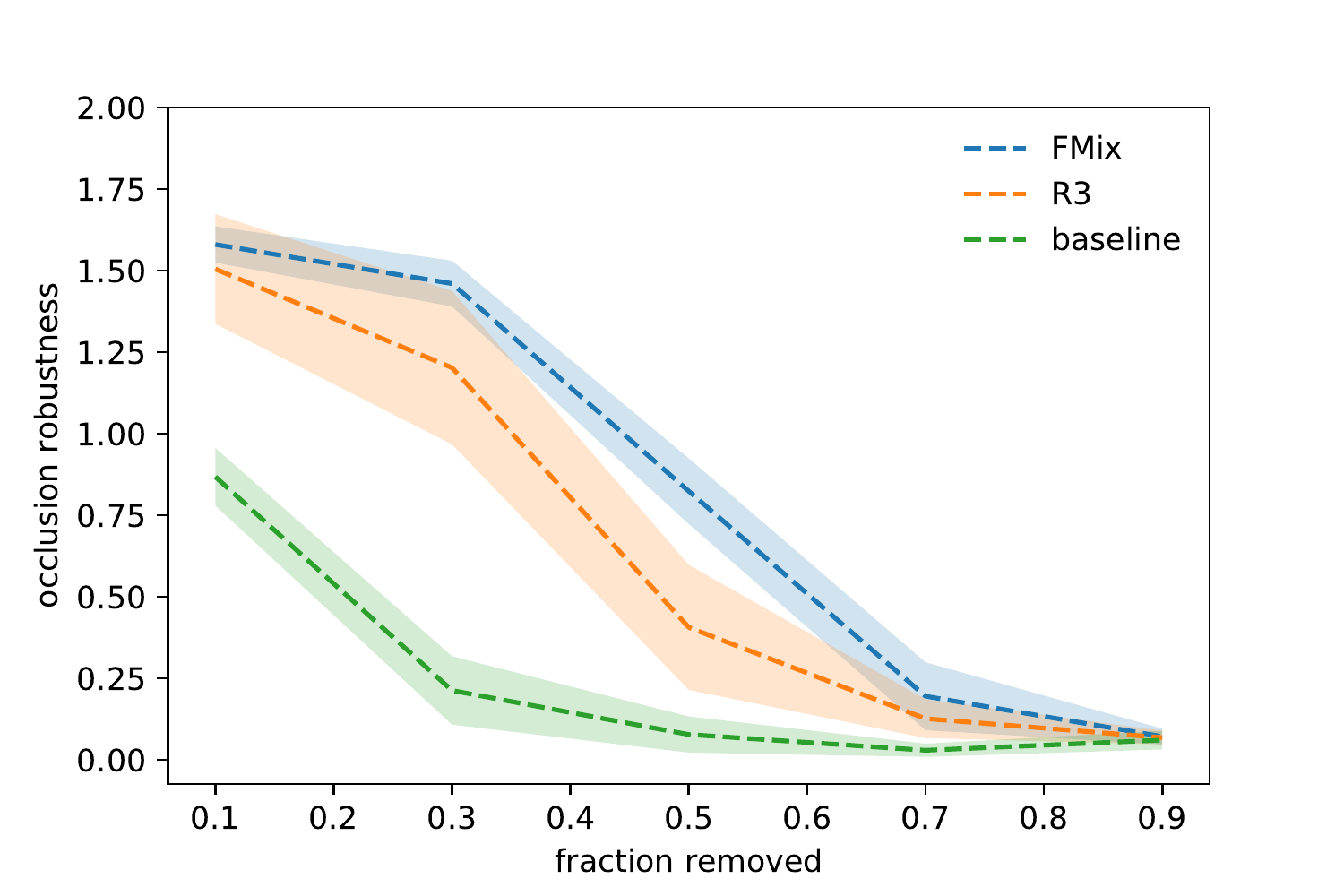}
    \end{subfigure}
    \caption{ \cutocc (left) and \iocc (right). Note that there is a difference in scale and the two should not be directly compared. We are rather interested in how the methods situate the different augmentation with respect to each other. It is important to notice that when measuring the robustness with \cutocc, RM3 appears significantly less robust than \fmix due to its sensitivity to patching with rectangles. On the other hand, \iocc highlights the robustness specific to \fmix.
    }
    \label{fig:3msks}
\end{figure*}

Figure~\ref{fig:3msks} also gives the results for \cutocc and \iocc for training with 3 random masks sampled from Fourier space. 

\begin{figure*}[]
    \centering
    \includegraphics[width=0.7\linewidth]{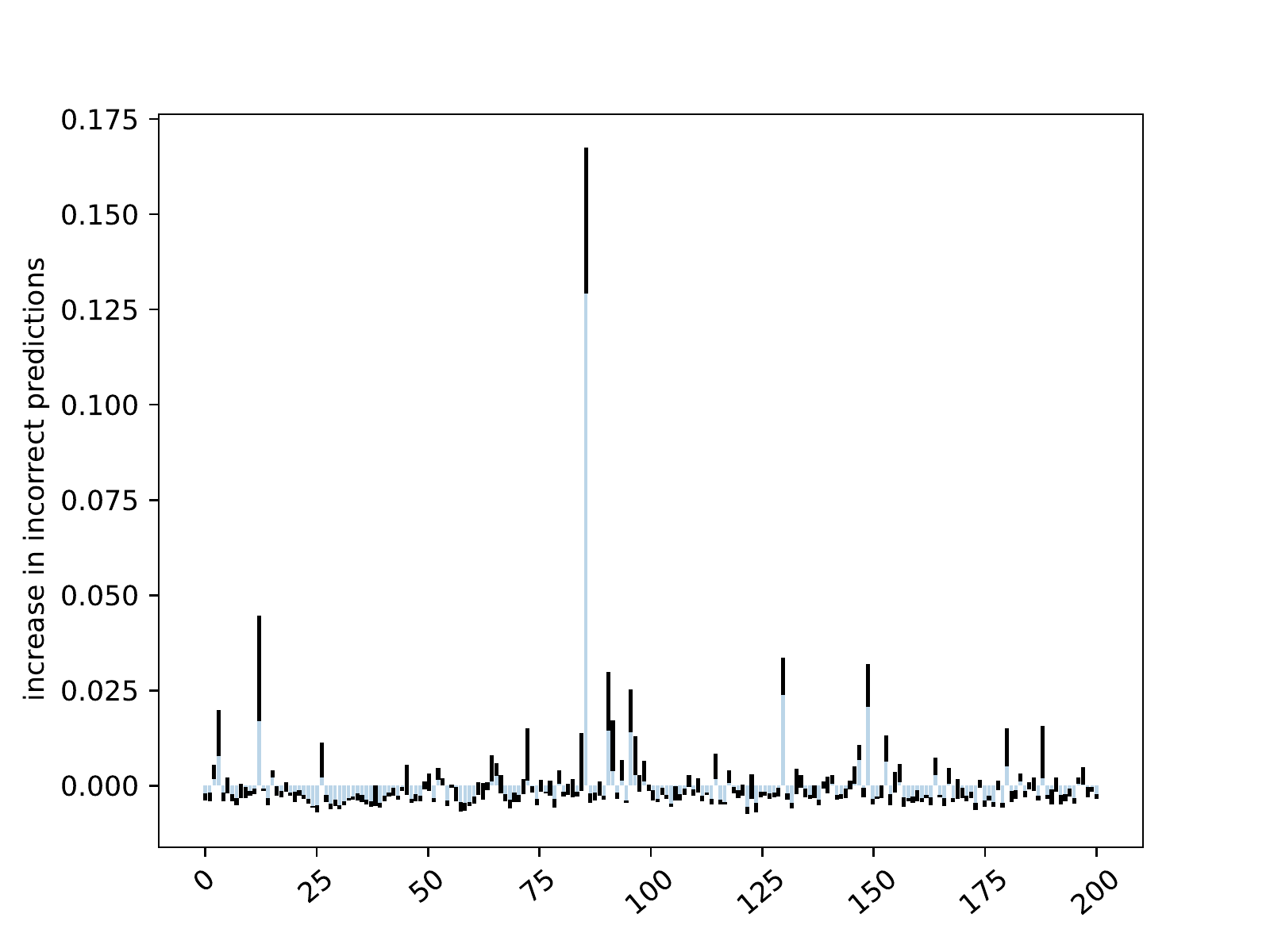}
    \caption{Difference between wrongly predicted classes when testing on original Tiny \imagenet data versus \cutout{} images.} 
    \label{fig:bag9_bias}
\end{figure*}

\subsubsection{BagNet shape and texture accuracy} \label{sup:tiny_jig}

We evaluate on the GST data set BagNet9 models trained on Tiny~\imagenet and present the results in Table~\ref{sup_tab:tiny_bag}. Despite the basic model displaying a bias towards predicting one of the classes when presented with patch-shuffled images (see Figure~\ref{fig:bag9_bias}), once again it does not have a lower texture or higher shape bias than the masked-based augmentations.

\section{Further results on \iocc experiments} \label{sup:further_results_iocc}
\subsection{Alternative \cutocc} \label{sup:unifrom}

Table~\ref{tab:beta21} gives the DI index when forcing the occluding patch to lie within image boundaries for patch sizes sampled from Beta(2,1). For the same scenario, but sampling uniformly from [0.1,\, 1] we present results in Table~\ref{tab:unif}.
Note that in the case of Tiny~\imagenet the bias is more visibly present for larger occluders. As such, uniformly sampling the patch size from the interval [0.3, 1] results in a DI index of $13.46_{\pm 5.74}$ for the basic model, while the level of data interference from \mixup is only $4.75_{\pm 1.93}$.
Similarly, for Fashion MNIST, when we increase the size of the occluder we obtain $0.65_{\pm0.20}$ for Mixup as opposed to $0.07_{\pm 0.11}$ for the basic model.
However, this does not change the conclusions of our experiments since, as mentioned in the main paper, robustness studies are usually carried out with large occluder sizes.

\begin{table}
        \centering
    \caption{DI index for sampling occluder size from a uniform distribution when the patch is restricted to lying within image boundaries and the size is sampled from Beta(2,1).} \label{tab:beta21}
    \begin{tabulary}{\linewidth}{lLLLL}
    \toprule
         & basic  & \mixup & \fmix   & \cutmix  \\
    \midrule
    \cifar{10}& $5.88_{\pm 1.82}$& $0.76_{\pm 0.69}$& $1.30_{\pm 1.27}$& $3.68_{\pm 3.66}$\\
    \cifar{100} & $29.71_{\pm 10.19}$& $6.80_{\pm 7.55}$& $6.08_{\pm 6.28}$& $13.19_{\pm 25.85}$\\
    Fashion MNIST & $0.67_{\pm 0.38}$& $3.51_{\pm 1.38}$& $1.87_{\pm 3.33}$& $1.78_{\pm 2.93}$\\
    Tiny~\imagenet & $15.25_{\pm 4.84}$& $6.38_{\pm 4.03}$& $5.87_{\pm 6.07}$& $13.97_{\pm 24.02}$\\
    \imagenet & $9.93$ & $28.72$ & $11.52$ & $-$\\
    \bottomrule
    \end{tabulary}
\end{table}

\begin{table}
        \centering
    \caption{DI index for sampling occluder size from a uniform distribution when the patch is restricted to lying within image boundaries and the size is sampled from [0.1, \,1] uniformly.} \label{tab:unif}
    \begin{tabulary}{\linewidth}{lLLLL}
    \toprule
         & basic  & \mixup & \fmix   & \cutmix  \\
    \midrule
    \cifar{10}& $5.74_{\pm 1.86}$ & $0.75_{\pm 0.69}$ &$1.25_{\pm 1.24}$ &$3.61_{\pm 3.60}$ \\
    \cifar{100} & $28.63_{\pm 9.85}$ &$6.31_{\pm 7.03}$ &
$5.86_{\pm 5.86}$ &$12.63_{\pm 24.69}$ \\
    Fashion MNIST & $1.88_{\pm 3.36}$ & $3.54_{\pm 1.33}$ &  $1.91_{\pm 3.33}$ & $1.76_{\pm 2.91}$ \\
    Tiny~\imagenet & $1.406_{\pm 04.36}$ & $0.581_{\pm 03.32}$ &
$0.553_{\pm 05.66}$ & $1.249_{\pm 21.44}$ \\
    \imagenet &$0.25$ & $0.48$ & $0.14$ &  $-$\\

    \bottomrule
    \end{tabulary}
\end{table}

\subsection{Occluding with images from another data set} \label{sup:mix_nomix}

\begin{figure*}
    \begin{subfigure}[t]{0.49\linewidth}
        \centering
        \includegraphics[width=\linewidth]{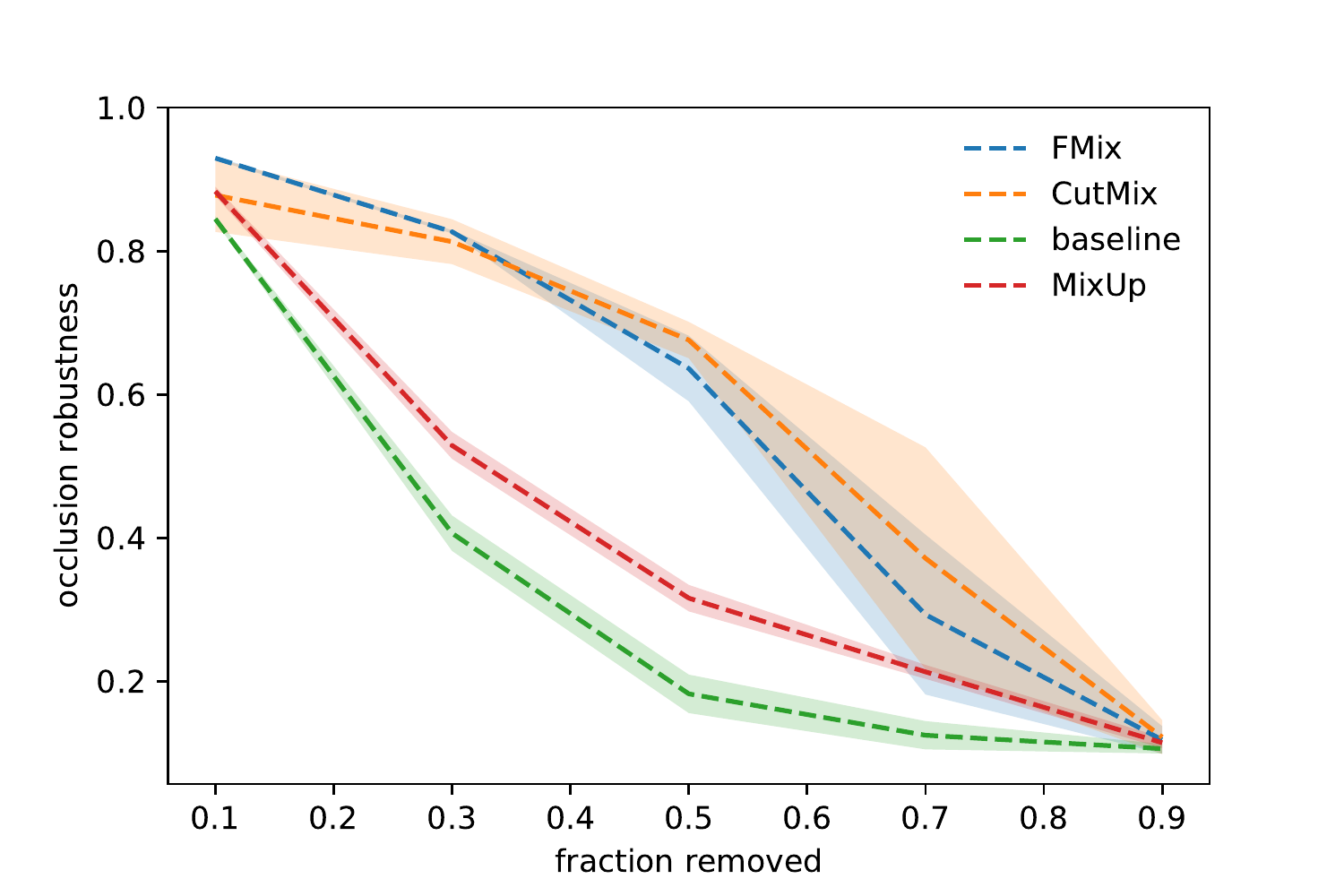}
        \caption{Uniform \cutocc}
    \end{subfigure}
    \hfill 
    \begin{subfigure}[t]{0.49\linewidth} 
        \centering
        \includegraphics[width=\linewidth]{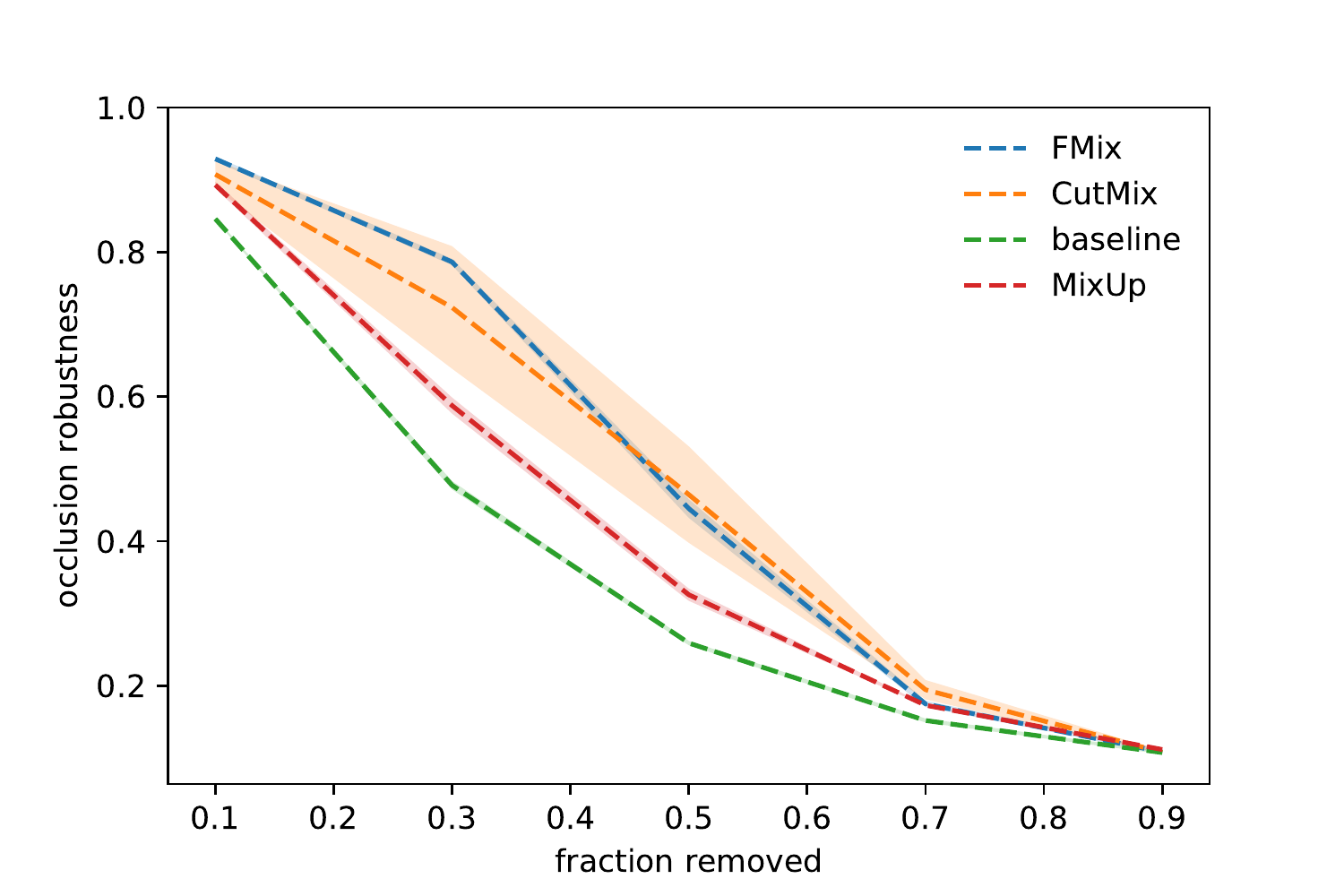}
        \caption{Non-uniform \cutocc} 
    \end{subfigure}
    \newline
    \begin{subfigure}[t]{0.49\linewidth} 
        \centering
        \includegraphics[width=\linewidth]{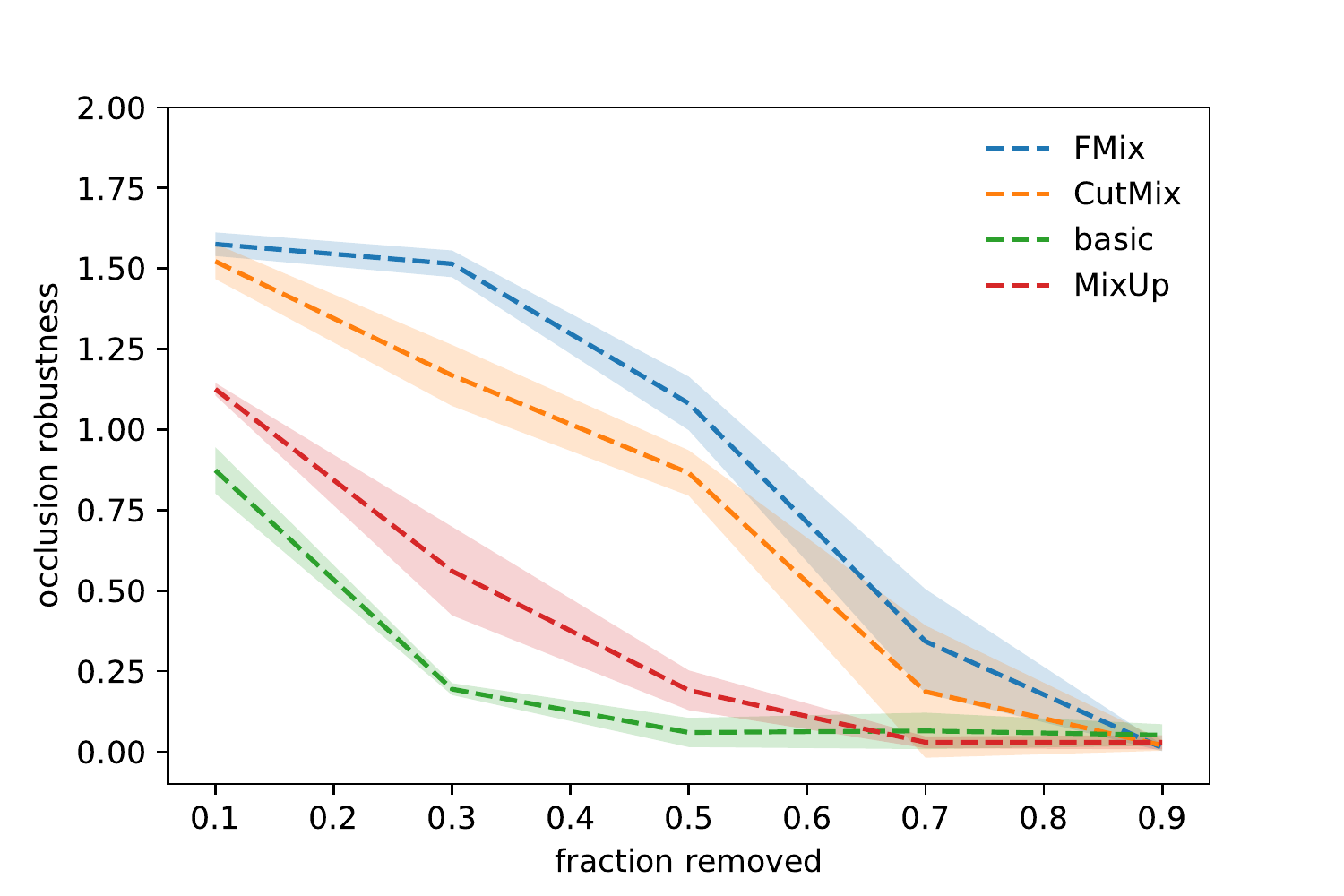}
        \caption{Uniform \iocc} 
    \end{subfigure}
    \hfill 
    \begin{subfigure}[t]{0.49\linewidth} 
        \centering
        \includegraphics[width=\linewidth]{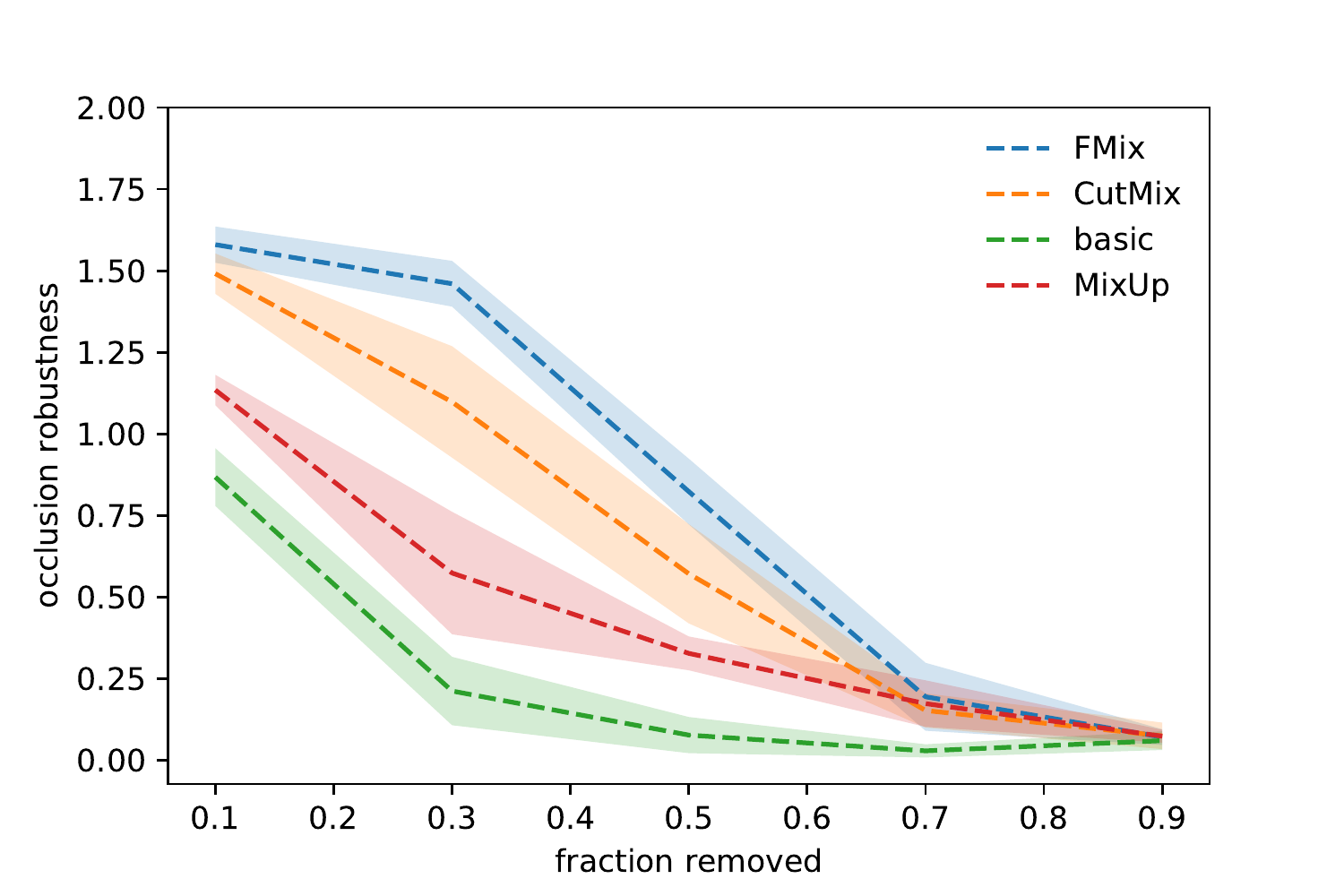}
        \caption{Non-uniform  \iocc} 
    \end{subfigure}
    \caption{Comparison of metric sensitivity to textured occlusion. Uniform occlusion refers to superimposing uniform patches over \cifar{10} images, while nonuniform refers to superimposing part of \cifar{100} samples. Nonuniform \cutocc provides significantly different results to its regular counterpart.}
    \label{fig:mix_nomix}
\end{figure*}

Since \cutocc does not account for the bias introduced by the occluding method, it is expected that changing the patch to a non-uniform one would greatly affect the results. For \cifar{10} models, Figure~\ref{fig:mix_nomix} presents the results of occluding with \cifar{100} images.
\iocc better rules out the specifics of the occluding patch, its uniform version giving similar results to the non-uniform one, whereas \cutocc pushes everything together.

\subsection{Randomising labels} \label{sup:rand_lbls}

To assess the sensitivity of \cutocc and \iocc to the overall performance of the model, we also experiment with randomising all the labels of the \cifar{10} data set.
When evaluated on the unaugmented training data, all the basic models achieve 100\% accuracy, while the \fmix models reach $99.99_{\pm0.01}$. 
Since all labels are corrupted, the accuracy on the test set before and after occlusion is no greater than random. 
However, the robustness of the augmentation-trained model can be seen on the training data, as captured by our metric (See Figure~\ref{fig:rand}).
On the other hand, \cutocc makes no distinction between learning with regular and augmented data (Table~\ref{tab:rand_oriz}).
Despite being such a peculiar case, it shows the comprehensiveness gained by accounting for the degradation on test data in relation to that on train.

\begin{table}[t]
    \begin{minipage}{0.55\linewidth}
    \centering
    \caption{Robustness to occluding with patches covering $50\%$ of each image. The models are trained with and without masking augmentation on data with randomised labels. 
    \cutocc makes no difference between regular and augmented training.} \label{tab:rand_oriz}
    \begin{tabulary}{\linewidth}{lLLL}
    \toprule
     & basic random & \fmix random & \fmix clean \\
     \midrule
    \cutocc &$10.24_{\pm0.27}$ &$9.78_{\pm0.18}$ &$63.63_{\pm4.54}$ \\
    \iocc   &$14.63_{\pm1.12}$ & $47.94_{\pm19.84}$ & $82.36_{\pm10.06}$\\
    \bottomrule
    \end{tabulary}
     \end{minipage}
     \hfill
    \begin{minipage}{0.4\linewidth}
        \centering
        \includegraphics[width=\linewidth]{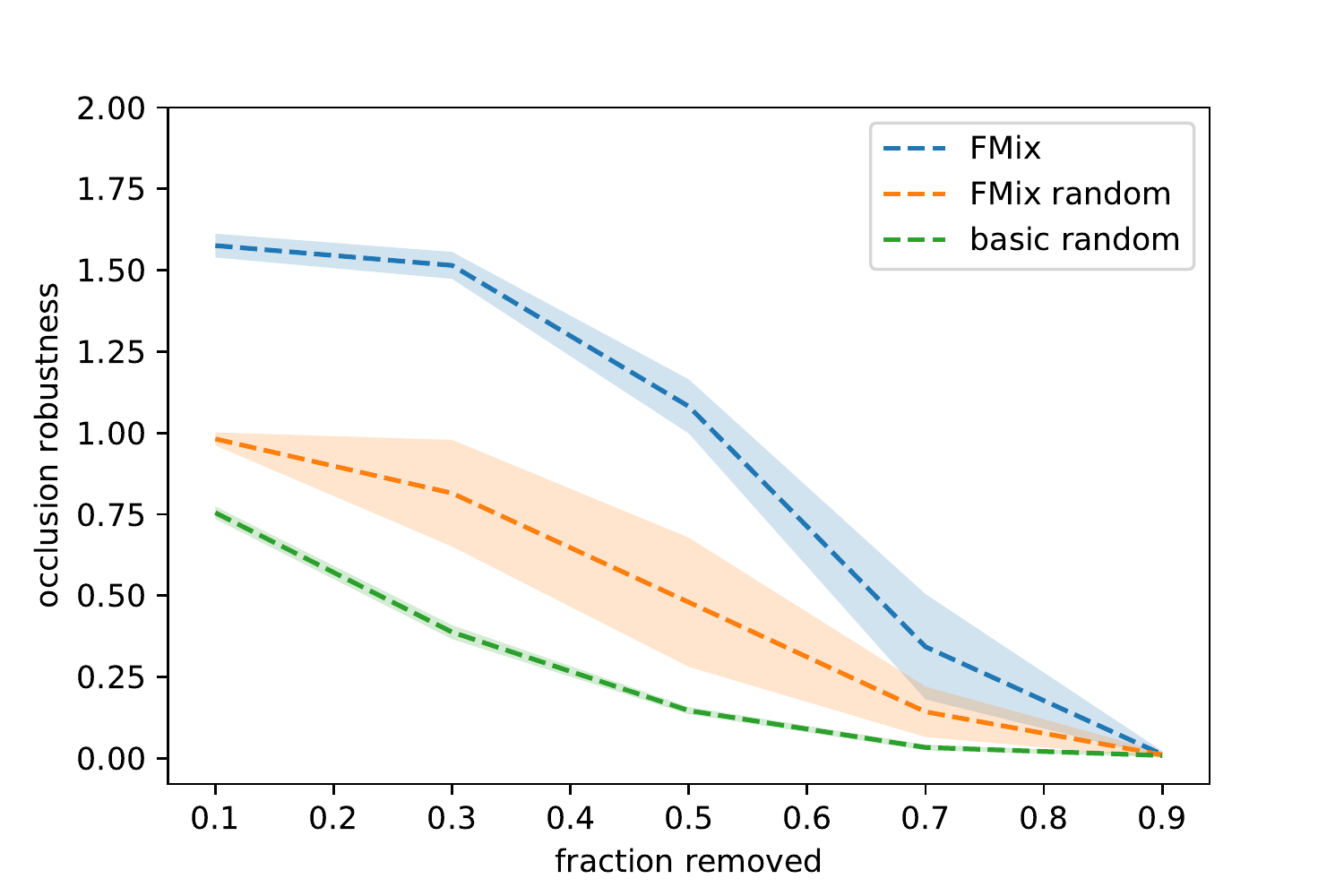}
         \captionof{figure}{\iocc results for training with clean and corrupted labels for basic and \fmix augmentation.} 
        \label{fig:rand}
    \end{minipage}

\end{table}

\subsection{Approximating \iocc} \label{sup:approximating}

\begin{figure*}
    \begin{subfigure}[t]{0.49\linewidth} 
        \centering
        \includegraphics[width=\linewidth]{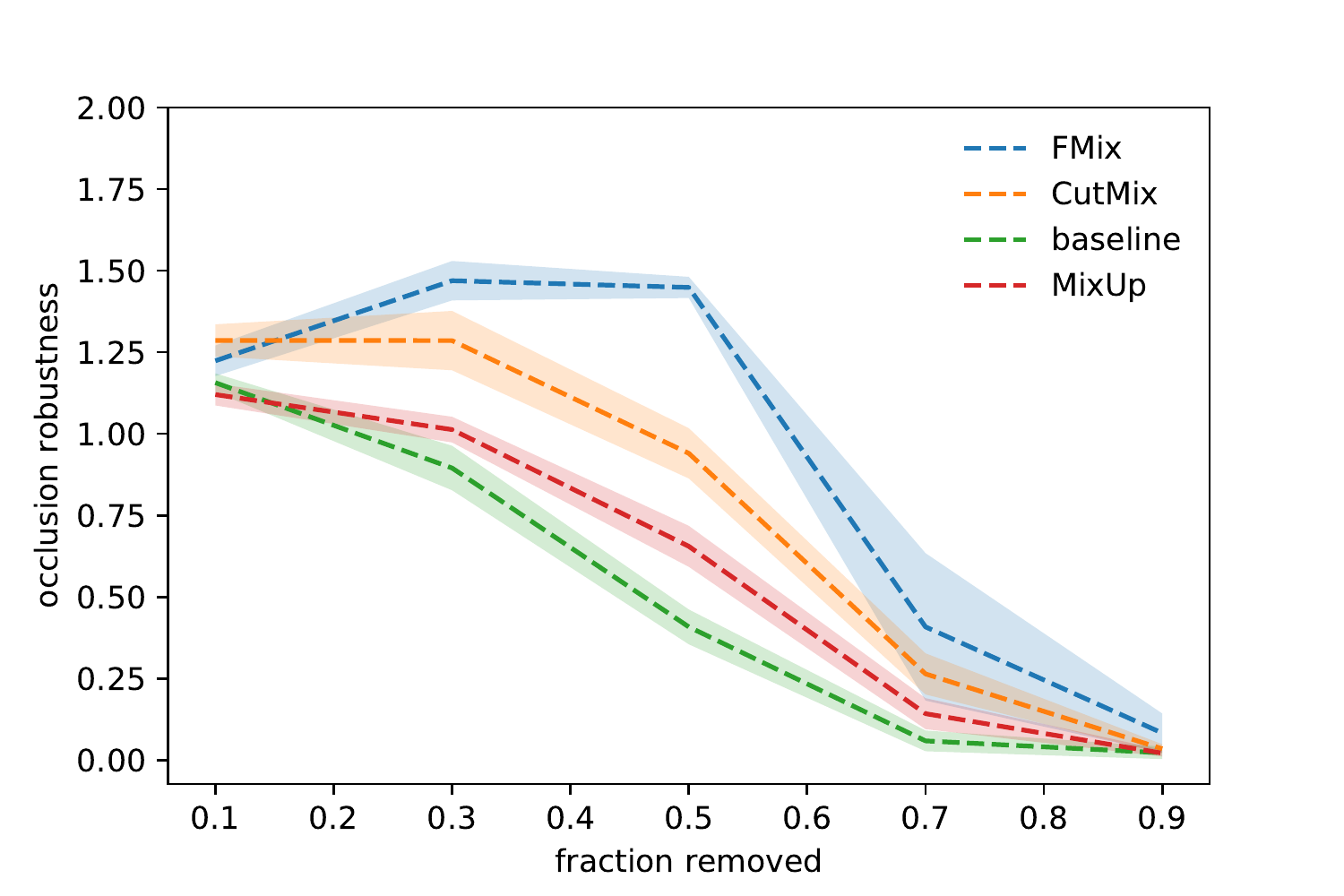}
        \caption{Fourier-sampled patches} 
    \end{subfigure}
    \hfill 
    \begin{subfigure}[t]{0.49\linewidth} 
        \centering
        \includegraphics[width=\linewidth]{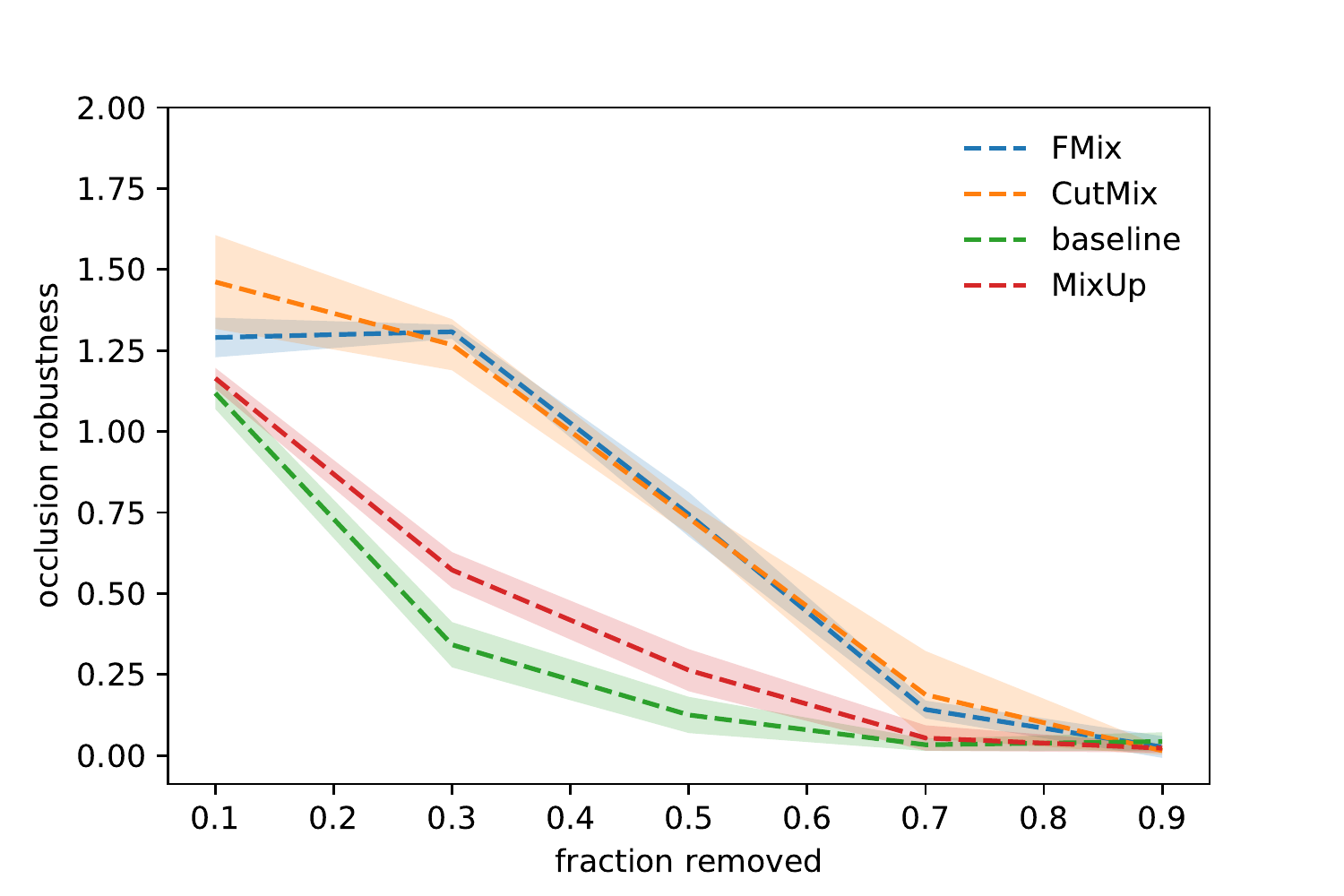}
        \caption{Rectangular patches} 
    \end{subfigure}
    \caption{Approximating \iocc with random masking can provide a first intuition, but is significantly noisier than using a saliency method.}
    \label{fig:approx}
\end{figure*}

As alternative methods for computing \iocc we experiment both with masks sampled from Fourier space and randomly positioned square patches.
Although using this type of random masking methods for computing $\mathcal{D}^{i}_{train}$ and $\mathcal{D}^{i}_{test}$ in Equation~(1) gives less precise results, it has the advantage of incurring less computation and can be used for rapid model analysis. 
For assessing a model across 5 runs for 6 different levels of occlusion, this method leads to a carbon footprint of 0.05 kgCO$_2$eq as opposed to 1.04 using Grad-CAM.
In Figure~\ref{fig:approx} we present the results obtained with these alternatives. 
We expect both methods to provide overoptimistic results for small patches, while Fourier sampling is expected to give more truthful scores as the size of the patch increases.
On the other hand, the contiguity of \cutout-based occlusion comes at the cost of not determining the robustness to multiple simultaneous occluders. 
This seems to play a role especially in the case of \cutmix augmentation.
Indeed, when superimposing a rectangular patch, it is difficult to differentiate \cutmix from \fmix-trained models.
To confirm that this is caused by the granularity of the occluders and not the shape, we also experiment with occluding using multiple rectangular patches.
We split the images in a $4 \times 4$ grid and occlude i\% of the tiles, obtaining results that are more similar to those obtained when occluding with Fourier-sample patches.
Thus, while significantly noisier, using randomly positioned occluders can provide an alternative for computing \iocc given that one takes into account the number of occluders.

\section{Removing the dominant class} \label{sup:removeDominnat}

\begin{figure*}
    \begin{subfigure}[t]{0.49\linewidth} 
        \centering
        \includegraphics[width=\linewidth]{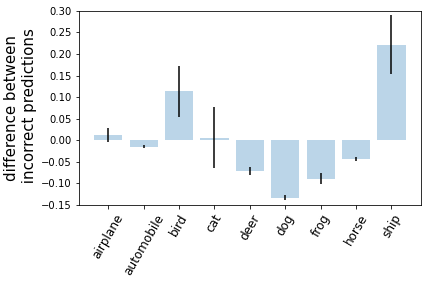}
        \caption{Fourier-sampled patches} 
    \end{subfigure}
    \hfill 
    \begin{subfigure}[t]{0.49\linewidth} 
        \centering
        \includegraphics[width=\linewidth]{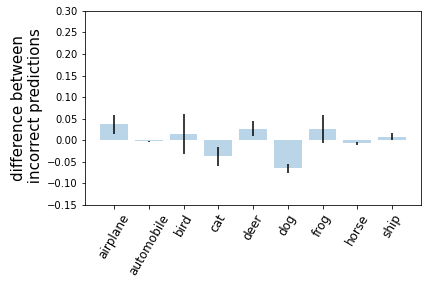}
        \caption{Rectangular patches} 
    \end{subfigure}
    \caption{Difference in incorrect predictions for the basic (left) and CutOut(right) models.}
    \label{fig:dominant}
\end{figure*}

We remove the 10th class from the \cifar{10} data set and retrain on the remaining classes.
In the main paper we give the results for occluding with non-uniform patches. 
When using black patches to obstruct images, we again identify a gap, but this time with respect to a CutOut-trained model (see Figure~\ref{fig:dominant}).
The basic model has a DI index of $1.23_{\pm 0.72}$, while CutOut $0.13_{\pm0.10}$. 
Thus, in both cases a model that is less affected by the artefacts than the basic model can be found.
Thus, when measuring \cutocc, the basic model will be disadvantaged.

\end{document}